\pdfoutput=1

\documentclass[11pt]{article}

\usepackage[final]{ACL2023}
\usepackage{times}
\usepackage{latexsym}

\usepackage[T1]{fontenc}

\usepackage[utf8]{inputenc}

\usepackage{microtype}

\usepackage{inconsolata}

\usepackage{comment}
\usepackage{graphicx}
\usepackage{amssymb} 
\usepackage{svg}
\usepackage{subcaption}
\usepackage{amsmath}
\usepackage{booktabs}
\usepackage{tabularx}
\usepackage{siunitx} 
\usepackage{subscript}
\usepackage{pifont}
\usepackage{xcolor} 
\usepackage{tikz}
\usepackage{multirow}
\usepackage{float}  
\usepackage{colortbl}
\usepackage{enumitem} 
\usepackage{latexsym}
\usepackage{amssymb}
\usepackage{arydshln}
\usepackage{makecell}
\newcolumntype{C}[1]{>{\centering\arraybackslash}p{#1}}

\definecolor{deepgreen}{rgb}{0.0, 0.5, 0.0} 
\definecolor{customblue}{HTML}{2F5597}

\title{DS$^2$-ABSA: Dual-Stream Data Synthesis with Label Refinement for Few-Shot Aspect-Based Sentiment Analysis}

\author{Hongling Xu$^{1,3}$, Yice Zhang$^{1,3}$, Qianlong Wang$^{1,3}$, Ruifeng Xu$^{1,2,3}$\thanks{\quad Corresponding Author}
\\
    $^{1}$ Harbin Institute of Technology, Shenzhen, China \\
    $^{2}$ Peng Cheng Laboratory, Shenzhen, China \\
    $^{3}$ Guangdong Provincial Key Laboratory of Novel Security Intelligence Technologies \\
     \texttt{xuhongling@stu.hit.edu.cn}, ~\texttt{xuruifeng@hit.edu.cn} \\
}

\begin{document}
\maketitle
\begin{abstract}


Recently developed large language models (LLMs) have presented promising new avenues to address data scarcity in low-resource scenarios. In few-shot aspect-based sentiment analysis (ABSA), previous efforts have explored data augmentation techniques, which prompt LLMs to generate new samples by modifying existing ones. However, these methods fail to produce adequately diverse data, impairing their effectiveness. Besides, some studies apply in-context learning for ABSA by using specific instructions and a few selected examples as prompts. Though promising, LLMs often yield labels that deviate from task requirements. To overcome these limitations, we propose DS$^2$-ABSA, a dual-stream data synthesis framework targeted for few-shot ABSA. It leverages LLMs to synthesize data from two complementary perspectives: \textit{key-point-driven} and \textit{instance-driven}, which effectively generate diverse and high-quality ABSA samples in low-resource settings. Furthermore, a \textit{label refinement} module is integrated to improve the synthetic labels. Extensive experiments demonstrate that DS$^2$-ABSA significantly outperforms previous few-shot ABSA solutions and other LLM-oriented data generation methods.\footnote{Our code and synthetic data are available publicly at \url{https://github.com/behappyplz/DS2-ABSA}}


\end{abstract}



\section{Introduction}

Aspect-based sentiment analysis (ABSA) aims to identify aspect terms and determine their sentiments within user reviews \cite{pontiki-etal-2014-semeval}.
For example, given the review ``\textit{the battery life is great, but the screen resolution is disappointing},'' the output of End-to-End ABSA (E2E-ABSA) would be \{(\textit{battery life}, {positive}), (\textit{screen resolution}, {negative})\}. Previous studies have proposed various deep learning methods~\cite{ijcai2022p0572,tian-etal-2023-end,scaria2024instructabsa}, demonstrating strong performance when trained on extensive manually labeled data. However, annotating sufficient data is extremely time-consuming and labor-intensive in practice, prompting researchers to explore ABSA approaches in low-resource scenarios \cite{varia-etal-2023-instruction,wang2023simple,wang2023chatgpt,zhang2024sentiment}.

Existing low-resource solutions comprise three main types. 
The first, \textbf{data augmentation}, produces additional samples by modifying existing ones, which are typically implemented through masked language modeling~\cite{li2020conditional, zhou-etal-2022-melm} or requiring large language models (LLMs)~\cite{dai2023chataug, peng-etal-2024-controllable}.
Although these methods can yield a large number of new samples, the diversity of them remains limited, thereby providing marginal benefit for subsequent model training.
The second type, \textbf{in-context learning}, aligns LLMs with the ABSA task through task-specific instructions and a few examples \cite{zhang2024sentiment,wang2024context,wang2023chatgpt}. 
However, even with well-crafted instructions and carefully chosen demonstrations, LLMs tend to deviate from task-specific requirements, frequently generating reasonable, yet incorrect results.
Apart from these two types, some researchers explore \textbf{pre-training} techniques 
\cite{wang2023simple,zhang2023empirical} to reduce reliance on downstream datasets. Nonetheless, these techniques require vast additional corpora and incur high training costs. 

Inspired by the recent advances in data synthesis~\cite{li-etal-2023-synthetic, wang-etal-2023-self-instruct}, we propose \textbf{DS$^2$-ABSA}, a dual-stream data synthesis framework for few-shot ABSA. Unlike existing methods, our study employs multi-granularity-guided synthesis targeted to E2E-ABSA.
Specifically, we first leverage LLMs to generate data via two distinct strategies: \textit{key-point-driven} and \textit{instance-driven}. 
The former engages LLMs to brainstorm a variety of potential ABSA attributes, which are then composed to create new samples. The latter transforms existing samples through operations including sample combination and selective reconstruction.
These two strategies are complementary: the former synthesizes data that covers a broader range of review scenarios and offers greater diversity, while the latter generates data based on existing samples and provides better relevance and quality.
Moreover, we integrate a \textit{label refinement} module to enhance the quality of labels in the synthetic data. 
This module applies a label normalization process alongside a noisy self-training algorithm that employs a few gold samples to guide the re-estimation of the synthetic labels.





Compared to previous methods, our approach offers the following advantages.
Firstly, by leveraging key-point-driven and instance-driven strategies, it can generate more diverse data than data augmentation methods, providing greater potential benefits for subsequent model training.
Secondly, in contrast to in-context learning methods, our approach introduces a novel way of utilizing LLMs, resulting in better task-specific alignment. The label refinement module further enhances this advantage.
Thirdly, unlike pre-training methods, our approach requires no additional corpus, entailing significantly lower training costs and avoiding the potential challenges of data acquisition.

Our contributions are summarized as follows:
\begin{enumerate}[itemsep=1.15pt, topsep=2.5pt]
    \item[(1)] We propose a dual-stream synthesis framework that leverages LLMs to generate ABSA samples from two distinct perspectives. To the best of our knowledge, this is the first exploration of data synthesis for E2E-ABSA.
    \item[(2)] We develop a label refinement module that effectively enhances the label quality of the synthesized data through label normalization and noisy self-training.
    \item[(3)] Experimental results on four public datasets demonstrate that DS$^2$-ABSA significantly outperforms existing low-resource ABSA solutions, as well as other LLM-based data augmentation and synthesis methods.
\end{enumerate}

\begin{figure*}[t]
    \centering
    \includegraphics[width=0.96\textwidth]{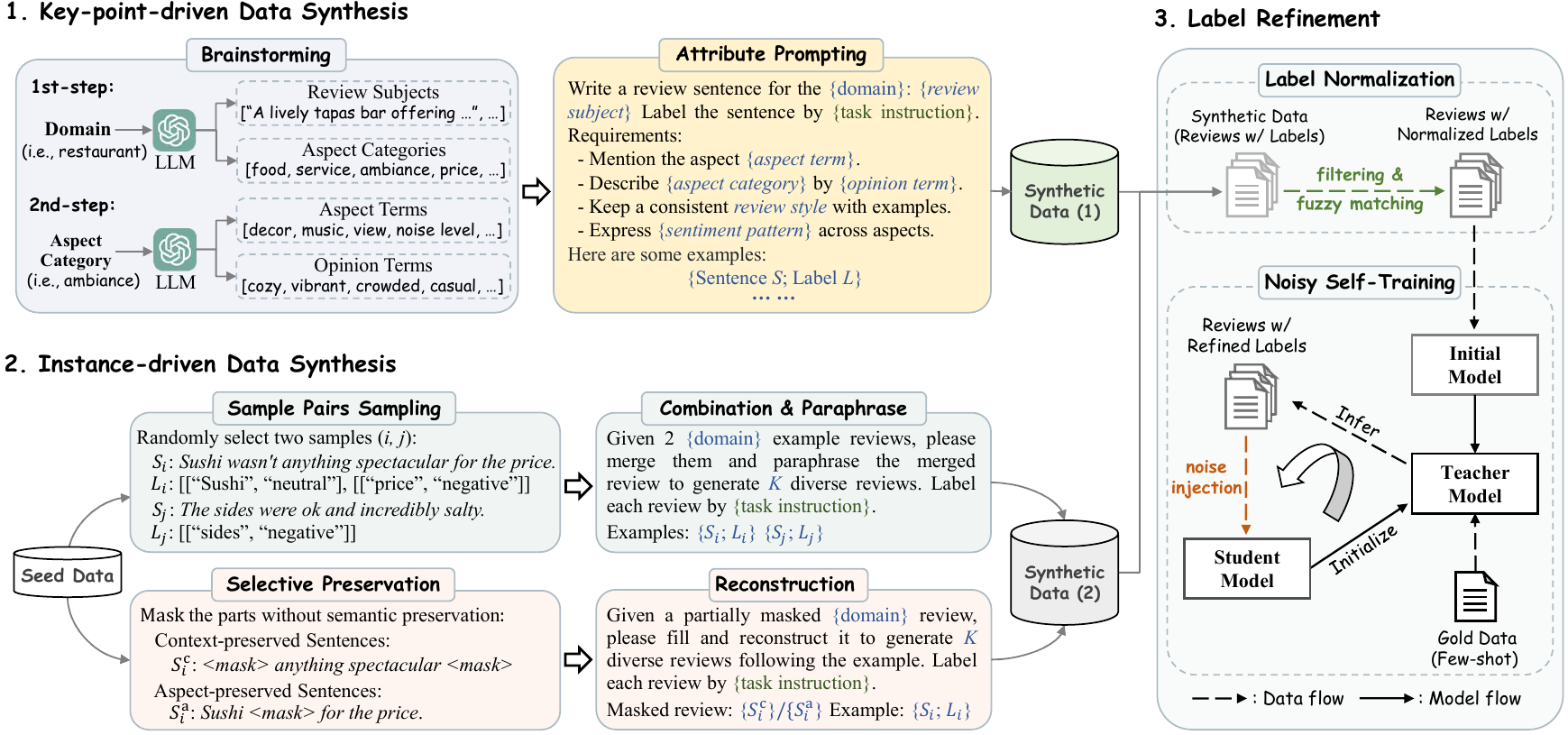}
    \caption{Overview of the proposed DS$^2$-ABSA. The process begins with parallel dual-stream data synthesis: the key-point-driven stream leverages LLMs to brainstorm a set of critical ABSA attributes for conditional generation, while the instance-driven stream applies a small seed dataset to perform multi-level transformations. The resulting data are then combined and processed through normalization and self-training for noise handling.}
    \label{fig: overview}
\end{figure*}

\section{Related Work}
\subsection{Few-shot ABSA}
Current state-of-the-art E2E-ABSA methods primarily rely on fine-tuning text-to-text language models \cite{zhang-etal-2021-aspect-sentiment,yu2023syngen,scaria2024instructabsa} or incorporating syntactic knowledge \cite{ijcai2022p0572,tian2023end,liu-etal-2024-lets} using sufficient labeled data. However, annotating fine-grained sentiments in adequate reviews is expensive. To mitigate data dependency, early methods mainly adopt data augmentation~\cite{wei2019eda, li2020conditional, ding-etal-2020-daga, hsu-etal-2021-semantics, zhou-etal-2022-melm} and pre-training~\cite{xu2019bert, zhou2020sentix, liu2023unified, wang2023simple, zhang2023empirical}.

Recently, the advent of powerful LLMs has inspired new approaches for few-shot ABSA, including (1) LLM-based data augmentation~\cite{dai2023chataug, peng-etal-2024-controllable}, which often lacks diversity in low-resource settings; (2) in-context learning~\cite{wang2023chatgpt, zhang2024sentiment, wang2024context, 10.1145/3626772.3657932, zhu-etal-2024-zzu}, where LLMs often fail to produce task-aligned outputs; and (3) knowledge distillation~\cite{zhou2024universalner}, which similarly requires substantial extra review corpora like pre-training, posing challenges in domains with limited data availability and privacy concerns. 
To tackle these issues, we propose a novel approach that utilizes LLMs to synthesize samples with rich diversity and high quality, whose labels are then refined to facilitate downstream training.

\subsection{Data Synthesis}
Data synthesis is a pivotal strategy to address challenges like high annotation costs and privacy issues \cite{long-etal-2024-llms}. With the advent of LLMs, it has been applied to various fields, such as text classification \cite{ye-etal-2022-zerogen, li-etal-2023-synthetic}, instruction tuning \cite{wang-etal-2023-self-instruct, zhao2024selfguide}, mathematical reasoning \cite{huang2024key, yu2024metamath, chan2024scaling}. 

Existing works can be grouped into three types: (1) instance-driven synthesis \cite{zhao2024selfguide, yu2024metamath}, which leverages examples to guide LLMS in synthesizing relevant data; (2) key-point-driven synthesis \cite{huang2024key, wang-etal-2024-improving-text}, utilizing conditional prompts to produce data satisfying specific attributes; and (3) knowledge-driven synthesis \cite{xu-etal-2024-knowledge, chan2024scaling}, incorporating external knowledge to steer the synthesis. In practice, these strategies are often combined to achieve optimal results. After generation, techniques such as data filtering~\cite{wang-etal-2023-self-instruct} are used for data curation. On this basis, our study pioneers the data synthesis for E2E-ABSA by designing key-point-driven and instance-driven strategies, along with a refinement module for label re-estimation.

\section{Method}
As depicted in Figure~\ref{fig: overview}, we propose a novel data synthesis framework to improve the few-shot E2E-ABSA,  Here, we detail our DS$^2$-ABSA pipeline, where \textit{key-point-driven} and \textit{instance-driven} data synthesis generate complementary ABSA samples from different perspectives. These synthetic data are merged and fed into the label refinement module to enhance label quality. 



\subsection{Key-point-driven Data Synthesis} 



This module aims to generate reviews based on key points, hereafter referred to as attributes. To this end, we define several critical attributes, such as aspect and opinion terms, and guide LLMs to \textit{brainstorm} numerous candidates for each attribute. Afterward, we sample a set of attributes from these candidates and employ LLMs to generate reviews based on these attributes using \textit{attribute prompting}.


\paragraph{Brainstorming.} 
Building on \citet{zhang2022survey}, we define the four core attributes that form a review sample: (a) \textit{review subject}, a general description of the restaurant or product, 
such as ``\textit{a lively tapas bar offering ...}'';
(b) \textit{aspect category}, indicating the generalized dimension being evaluated, such as `\textit{ambiance}' and `\textit{service}';
(c) \textit{aspect term}, referring to the specific opinion target explicitly mentioned in the review, such as `\textit{decor}' and `\textit{noise}'; and (d) \textit{opinion term}, representing the descriptive expression conveying sentiment to the opinion target, such as `\textit{charming}' and `\textit{cozy}'.


Next, we implement a coarse-to-fine generation strategy to produce a range of potential values for each attribute. Firstly, given a specific domain (such as restaurants or laptops), we prompt LLMs to brainstorm and generate representative review subjects and aspect categories. Secondly, for each aspect category generated, we guide LLMs to generate a diverse array of aspects and opinion terms. We then collect the values for each attribute, forming the corresponding candidate pools.
Following \citet{wang-etal-2024-improving-text}, we employ GPT-4~\cite{openai2023gpt4} for brainstorming to ensure both the quantity and quality of the attribute candidates.

\paragraph{Attribute Prompting.}


The attributed generation consists of three steps. Firstly, we randomly sample a set of attributes from the brainstormed candidate pools, denoted as ($rs_i$, $ac_i$, $at_i$, $ot_i$). Secondly, we instruct LLMs to generate a review that concerns the review subject $rs_i$ and includes the aspect category $ac_i$, aspect term $at_i$, and opinion term $ot_i$. It is important to note that we do not sample multiple sets of attributes and combine them to generate reviews, as this method could lead to potential conflicts among different attribute sets and reduce the coherence of the generated reviews. Finally, we require LLMs to generate the corresponding ABSA labels based on the provided attributes and the generated reviews.


%


Our observations indicate that, despite utilizing diverse attributes, the generated reviews tend to exhibit a uniform style and express sentiments too simplistically, thereby deviating from the true data distribution. To address these issues, we introduce two control attributes in our prompts: \textit{review style} and \textit{sentiment pattern}. The review style attribute involves using a few real reviews as exemplars to guide LLMs in generating reviews that mimic a similar style. The sentiment pattern attribute dictates the method of expressing sentiments, with options including `consistent,' `mixed,' and `implicit.' This allows for control over how reviews express sentiments—whether consistently across different aspects, in a varied manner, or implicitly, where sentiments are conveyed indirectly through context rather than explicit opinion words. The full prompt is presented in Appendix \ref{sec: prompt}.

\subsection{Instance-driven Data Synthesis}


This module synthesizes new data by transforming existing data, differing from the key-point-driven module that synthesizes data from scratch. The main advantage of this approach is that the synthesized data maintains strong in-domain relevance with the reference samples. More importantly, during the data synthesis process, LLMs can access the labels of the reference samples, thereby ensuring a higher quality of the synthesized labels. Specifically, we employ two 
operations to facilitate this transformation: \textit{sample combination} and \textit{selective reconstruction}.


\paragraph{Sample Combination.}
This operation randomly selects two samples from the seed data and instructs LLMs to merge them, thereby creating a new sample. Such a combination can effectively increase sample diversity, as it simulates a broader range of review scenarios. However, a potential issue with this approach is that this may lead to semantic discontinuities and content conflicts. To address this, we additionally require LLMs to paraphrase the merged samples, aiming to produce more coherent and consistent samples.
\paragraph{Selective Reconstruction.} 
This operation is inspired by existing data augmentation methods \cite{wei2019eda, li2020conditional, hsu-etal-2021-semantics}. It begins by preserving a portion of segments in a given review and then directs LLMs to reconstruct the complete review. We develop two selective preservation strategies: \textit{context preservation} and \textit{aspect preservation}. Context preservation masks the aspect terms and their surrounding $m$ words. 
Aspect preservation randomly masks segments of the given review except for the aspect terms, with a total masking portion of $p_{mask}$. Subsequently, these masked reviews are input into LLMs, tasked with generating complete reviews. Compared to traditional data augmentation methods, the advantage of this operation is that it preserves fewer review segments and leverages the capabilities of LLMs to generate more diverse samples.

\subsection{Label Refinement}
The ABSA data synthesized by LLMs inevitably contain inaccurate labels due to misalignment with task requirements~\cite{wang2023chatgpt}. To reduce their impact, our study introduces a novel label re-estimation method that rectifies erroneous labels by \textit{label normalization} and \textit{noisy self-training}.

\paragraph{Label Normalization.}  
For initial refinement, we introduce a rule-based approach to normalize the synthetic labels. In the task-specific requirements of ABSA, aspect terms must appear explicitly as complete sub-sequences within the text. By leveraging this rule, we compare the extracted aspects in synthetic labels with their corresponding sentences in the synthetic data, removing any that do not appear as matches. We then apply fuzzy matching for matched but incomplete aspects, substituting them with n-grams that minimize the Levenshtein distance per unit length.

\paragraph{Noisy Self-training.} 
We implement the noisy self-training algorithm~\cite{Xie_2020_CVPR, liu-etal-2021-noisy, jiang-etal-2023-noisy} to re-estimate the synthetic labels using a few gold data. The process begins by training an initial model on normalized data, followed by fine-tuning with gold data to obtain the teacher model $\mathcal{T}_0$, which is expected to better align with task requirements.

Next, we iterate through the noisy student training process. In the $i$-th iteration, we first use the previous teacher model $\mathcal{T}_{i-1}$ to label the synthetic sentences, refining errors like aspect boundary inaccuracies and sentiment misinterpretations. A new student model $\mathcal{S}_{i}$ is then trained on the refined data, with noise injected by randomly deleting or masking tokens in 50\% of samples at a disturbance probability $p_{\text{noise}}$ to improve robustness. $\mathcal{S}_{i}$ is subsequently fine-tuned on a few gold data to produce the updated teacher model $\mathcal{T}_{i}$. The iterative process repeats until the performance stabilizes, after which the final teacher model serves as the ABSA model for evaluation.

\section{Experiments}

\subsection{Settings}
\paragraph{Datasets.} We evaluate the proposed method on four ABSA benchmark datasets, including Lap14 and Res14 from \citet{pontiki-etal-2014-semeval}, Res15 from \citet{pontiki-etal-2015-semeval}, and Res16 from \citet{pontiki-etal-2016-semeval}. These datasets cover two domains: \textit{restaurant} and \textit{laptop}. The data statistics are presented in Table~\ref{tab: stat}, where we randomly split 20\% of the training set for validation. 

\begin{table}[h]
    \centering
    \footnotesize
    \renewcommand{\arraystretch}{0.75} 
        \begin{tabular}{@{}p{1cm}@{}p{0.65cm}@{}!{\vrule width 0.5pt}C{1.2cm}@{}C{1.2cm}@{}C{0.85cm}@{}C{0.85cm}@{}C{0.85cm}@{}C{0.85cm}@{}}
        \toprule
        \multicolumn{2}{c!{\vrule width 0.5pt}}{\textbf{Dataset}}  & \textbf{Samples} & \textbf{Aspects} & \textbf{\#Pos} & \textbf{\#Neu} & \textbf{\#Neg} & \textbf{\#Con}  \\

            \midrule
            \multirow{3}{*}{Lap14} & train & 2,436   & 1,922     & 808  & 387   & 691 &  36  \\
            & dev   & 609    & 436      & 179   & 73    & 175 &  9  \\
            & test  & 800    & 654      & 341   & 169    & 128 &  16 \\
            \cmidrule{1-8}
            \multirow{3}{*}{Res14} &train & 2,432    & 2,972     & 1,774   & 509   & 621 &  68 \\
            & dev   & 609    & 721      & 390   & 124    & 184 &  23 \\
            & test  & 800    & 1,134      & 728   & 196    & 196 &  14 \\
            \cmidrule{1-8}
            \multirow{3}{*}{Res15} & train & 1,052   & 956     & 721  & 29   & 199 &  7 \\
            & dev   & 263    & 243      & 181   & 5    & 53  &  4 \\
            & test  & 685    & 542      & 319   & 27    & 179 &  17 \\
            \cmidrule{1-8}
            \multirow{3}{*}{Res16} & train  & 1,600   & 1,363     & 952  & 51   & 337  & 23 \\
            & dev    & 400    & 380      & 268   & 10    & 96  &  6 \\
            & test   & 676    & 612      & 460   & 28    & 113  &  11 \\
        \bottomrule
        \end{tabular}
    \caption{Statistics of the four ABSA datasets. \#Pos, \#Neu, \#Neg, and \#Con represent positive, neutral, negative, and conflict aspects, respectively. }
    \label{tab: stat}
\end{table}

\paragraph{Implementation Details.}
In experiments, we adopt two few-shot settings: \textbf{2\%-shot} and \textbf{5\%-shot}, wherein a corresponding proportion of training data is randomly sampled to simulate low-resource scenarios. 
Unless specified otherwise, we employ \textbf{GPT-3.5 Turbo}~\cite{gpt35utrbo} for data generation.\footnote{The specific version we use is \textit{gpt-3.5-turbo-0125}.} 
Training samples with explicit aspects are selected as seed data. In key-point-driven synthesis, we generate 20,000 samples using randomly combined attributes and 4 examples in prompts. In instance-driven synthesis, we limit the maximum combined samples to 1,000, set the aspect masking window $m$ to 0 and 2, and the context masking probability $p_{mask}$ to 0.6 with random masking twice. 
Besides, the number of generated samples in a single response $K$ is 4. 
See Appendix~\ref{sec: prompt} for the detailed prompts applied in the dual-stream data synthesis. During the label refinement process, \(p_{\text{noise}}\) is set to 0.1, the maximum number of iterations is 3, and additional hyper-parameters can be found in Appendix~\ref{sec: para}. All experiments are conducted on NVIDIA A6000 GPUs. We run fine-tuning experiments with three random seeds and report the average \textbf{F1 score}. 

\begin{table*}[t]
    \centering
    \footnotesize
    \begin{tabular}{@{}C{2.9cm}@{}p{1.8cm}@{}C{1.0cm}@{}C{1.0cm}@{}C{1.0cm}@{}C{1.0cm}@{}C{1.65cm}@{}C{1.0cm}@{}@{}C{1.0cm}@{}@{}C{1.0cm}@{}@{}C{1.0cm}@{}C{1.65cm}@{}}
    \toprule
    \multirow{2.5}{*}{\textbf{ABSA Model}} & \multirow{2.5}{*}{\textbf{Method}} & \multicolumn{5}{c}{\textbf{2\%-shot}} & \multicolumn{5}{c}{\textbf{5\%-shot}} \\
    \cmidrule(lr){3-7} \cmidrule(lr){8-12}
    & & \textbf{Lap14} & \textbf{Res14}& \textbf{Res15} & \textbf{Res16} & \textbf{Avg($\Delta$)} & \textbf{Lap14} & \textbf{Res14} & \textbf{Res15} & \textbf{Res16} & \textbf{Avg($\Delta$)} \\
    \midrule
     \multirow{8.5}{*}{TAG\scalebox{1.0}[1.0]{-}BERT} & Origin & 15.83 & 37.49 & 23.04 & 20.19 & 24.14 & 35.29 & 51.64 & 34.52 & 43.48 & 41.23 \\
     \cdashline{2-12}
     \addlinespace[1pt]
    \multirow{8}{*}{\cite{hu-etal-2019-open}} 
    & MELM\textsuperscript{$\dagger$} & 38.27 & 46.26 & 32.11 & 34.90 & 37.89\textsubscript{+13.75} & 42.86 & 57.39 & 39.76 & 51.04 & 47.76\textsubscript{+6.53} \\
    & AugGPT\textsuperscript{$\dagger$} & 35.29 & 48.92 & 28.19 & 39.69 & 38.02\textsubscript{+13.88} & 37.26 & 57.93 & 37.88 & 53.26 & 46.58\textsubscript{+5.35} \\
    & CoTAM\textsuperscript{$\dagger$} & 39.15 & 56.05 & 31.06 & 42.36 & 42.16\textsubscript{+18.02} & 45.21 & 59.07 & 43.99 & 54.18 & 50.61\textsubscript{+9.38} \\
     \cdashline{2-12}
     \addlinespace[1pt]
    & BERT\scalebox{1.0}[1.0]{-}PT\textsuperscript{$\ddagger$} & 40.66 & 49.39 & 35.42 & 46.67 & 43.04\textsubscript{+18.90} & 47.95 & 61.92 & 35.75 & 54.19 & 49.95\textsubscript{+8.72} \\
    & SPT\scalebox{1.0}[1.0]{-}ABSA\textsuperscript{$\ddagger$} & 35.92 & 47.56 & 31.64 & 42.74 & 39.47\textsubscript{+15.33} & 46.71 & \textbf{63.58} & 40.42 & 55.17 & 51.47\textsubscript{+10.24} \\
     \cdashline{2-12}
     \addlinespace[1pt]
    & ZeroGen\textsuperscript{$\ast$} & 41.84 & 55.33 & 41.25 & 48.86 & 46.82\textsubscript{+22.68} & 45.87 & 56.64 & 41.92 & 53.77 & 49.55\textsubscript{+8.32} \\
    & Self\scalebox{1.0}[1.0]{-}Instruct\textsuperscript{$\ast$} & 41.85 & 56.94 & 42.34 & 52.71 & 48.46\textsubscript{+24.32} & 46.13 & 59.54 & 44.57 & 55.80 & 51.51\textsubscript{+10.28}  \\
    & \textbf{DS\textsuperscript{2}\scalebox{1.0}[1.0]{-}ABSA}\textsuperscript{$\ast$}  & \textbf{47.30} & \textbf{60.39} & \textbf{49.49} &\textbf{59.40} & \textbf{54.15\textsubscript{+30.01}} & \textbf{47.97} & 62.37 & \textbf{49.26} & \textbf{58.40} & \textbf{54.50\textsubscript{+13.27}} \\
    \midrule
    \multirow{8.5}{*}{\textsc{Paraphrase}} & 
    Origin & 47.64 & 53.40 & 39.23  & 39.75 & 45.01 & 51.51 & 62.01 & 51.45 & 52.58 & 54.39 \\
     \cdashline{2-12}
     \addlinespace[1pt]
    \multirow{8}{*}{\cite{zhang-etal-2021-aspect-sentiment}}  
    & MELM\textsuperscript{$\dagger$} & 48.57 & 57.11 & 41.84 & 48.74 & 49.07\textsubscript{+4.06} & 52.85 & 63.42 & 48.01 & 60.20 & 56.12\textsubscript{+1.73} \\
    & AugGPT\textsuperscript{$\dagger$} & 46.81 & 57.80 & 49.18 & 47.28 & 50.27\textsubscript{+5.26} & 48.74 & 62.97 & 52.71 & 59.13 & 55.89\textsubscript{+1.50} \\
    & CoTAM\textsuperscript{$\dagger$} & 47.28 & 58.71 & 46.75 & 53.15 & 51.47\textsubscript{+6.46} & 52.55 & 62.37 & 52.06 & 59.37 & 56.59\textsubscript{+2.20}  \\
    \cdashline{2-12}
    \addlinespace[1pt]
    & FS\scalebox{1.0}[1.0]{-}ABSA\textsuperscript{$\ddagger$} & 49.66 & 57.65 & 47.15 & 50.70 & 51.29\textsubscript{+6.28} & 52.02 & 63.12 & 53.30 & 61.52 & 57.49\textsubscript{+3.10} \\
    & UniNER\textsuperscript{$\ddagger$} & 53.92 & 64.02 & \textbf{53.76} & 58.36 & 57.52\textsubscript{+12.51} & 54.75 & 67.34 & \textbf{55.41} & 60.04 & 59.39\textsubscript{+5.00} \\
    \cdashline{2-12}
    \addlinespace[1pt]
    & ZeroGen\textsuperscript{$\ast$} & 48.85 & 57.71 & 49.41 & 55.10 & 52.77\textsubscript{+7.76} & 51.92 & 63.18 & 53.58 & 59.19 & 56.97\textsubscript{+2.58} 
  \\
    & Self\scalebox{1.0}[1.0]{-}Instruct\textsuperscript{$\ast$}  & 50.55 & 61.02 & 51.27 & 56.56 & 54.85\textsubscript{+9.84} & 53.67 & 64.89 & 51.82 & 60.36 & 57.69\textsubscript{+3.30}
 \\ 
    & \textbf{DS\textsuperscript{2}\scalebox{1.0}[1.0]{-}ABSA}\textsuperscript{$\ast$}  & \textbf{56.86} & \textbf{64.92} & 53.15 & \textbf{61.16} & \textbf{59.02\textsubscript{+14.01}} & \textbf{60.83} & \textbf{68.41} & 54.32 & \textbf{61.87} & \textbf{61.36\textsubscript{+6.97}} \\
    \midrule
    \multirow{8.5}{*}{\textsc{Instruct}ABSA} & Origin & 52.05 & 63.36 & 53.67 & 58.78 & 56.97 & 57.54 & 67.59 & 55.08 & 62.91 & 60.78 \\
\cdashline{2-12}
\addlinespace[1pt]
\multirow{8}{*}{\cite{scaria2024instructabsa}} 
& MELM\textsuperscript{$\dagger$} & 56.36 & 64.24 & 52.22 & 59.81 & 58.16\textsubscript{+1.19} & 58.53 & 68.15 & 54.31 & 63.44 & 61.11\textsubscript{+0.33} \\
& AugGPT\textsuperscript{$\dagger$} & 53.40 & 63.26 & 54.26 & 61.27 & 58.05\textsubscript{+1.08} & 55.11 & 66.54 & 57.25 & 64.93 & 60.96\textsubscript{+0.18} 
 \\
& CoTAM\textsuperscript{$\dagger$} & 54.07 & 64.77 & 51.42 & 62.45 & 58.18\textsubscript{+1.21} & 56.41 & 67.35 & 55.48 & 63.84 & 60.77\textsubscript{-0.01}
 \\
\cdashline{2-12}
\addlinespace[1pt]
& FS\scalebox{1.0}[1.0]{-}ABSA\textsuperscript{$\ddagger$} & 56.59 & 63.46 & 55.88 & 61.55 & 59.37\textsubscript{+2.40} & 61.63 & 68.68 & 55.83 & 66.24 & 63.10\textsubscript{+2.32} \\
& UniNER\textsuperscript{$\ddagger$} & 56.95 & 68.90 & 56.82 & 62.68 & 61.34\textsubscript{+4.37} & 59.81 & 70.33 & 58.34 & 67.95 & 64.11\textsubscript{+3.33}
 \\
\cdashline{2-12}
\addlinespace[1pt]
& ZeroGen\textsuperscript{$\ast$} & 54.72 & 64.22 & 53.85 & 62.56 & 58.84\textsubscript{+1.87} & 56.93 & 66.61 & 56.53 & 64.06 & 61.03\textsubscript{+0.25}
 \\
& Self\scalebox{1.0}[1.0]{-}Instruct\textsuperscript{$\ast$}  & 56.78 & 66.98 & 56.15 & 61.84 & 60.44\textsubscript{+3.47} & \textbf{61.89} & 68.84 & 57.39 & 67.07 & 63.80\textsubscript{+3.02} \\
& \textbf{DS\textsuperscript{2}\scalebox{1.0}[1.0]{-}ABSA}\textsuperscript{$\ast$}  & \textbf{58.15} & \textbf{69.65} & \textbf{59.78} & \textbf{63.89} & \textbf{62.87\textsubscript{+5.90}} & 61.24 & \textbf{71.94} & \textbf{60.56} & \textbf{68.81} & \textbf{65.64\textsubscript{+4.86}} \\
    \bottomrule
    \end{tabular}
    \caption{Main results compared with low-resource enhancement approaches (more baseline results are presented in Table~\ref{tab:more}). Data augmentation and data synthesis methods are marked with \textsuperscript{$\dagger$} and \textsuperscript{$\ast$}, respectively. Pre-training and distillation methods, marked with \textsuperscript{$\ddagger$}, rely on large amounts of additional unsupervised data.}
    \label{tab: slm}
\end{table*}

\begin{table*}[t]
    \centering
    \footnotesize
    \begin{tabular}{@{}C{2.7cm}@{}p{3.15cm}@{}C{1cm}@{}C{1cm}@{}C{1cm}@{}C{1cm}@{}C{1cm}@{}C{1cm}@{}C{1cm}@{}C{1cm}@{}C{1cm}@{}C{1cm}@{}}
    \toprule
    \multirow{2.5}{*}{\textbf{Backbone LLM}} & \multirow{2.5}{*}{\textbf{~~~~~~~~Method}} & \multicolumn{5}{c}{\textbf{2\%-shot}} & \multicolumn{5}{c}{\textbf{5\%-shot}} \\
    \cmidrule(lr){3-7} \cmidrule(lr){8-12}
    & & \textbf{Lap14} & \textbf{Res14}& \textbf{Res15} & \textbf{Res16} & \textbf{Avg} & \textbf{Lap14} & \textbf{Res14} & \textbf{Res15} & \textbf{Res16} & \textbf{Avg} \\
    \midrule
    \multirow{2}{*}{GPT-4} & Zero-shot Prompting & 41.35 & 60.61 & 49.42 & 53.25 & 51.16 & 41.35 & 60.61 & 49.42 & 53.25 & 51.16 \\
    & In-context Learning & 45.24 & 65.10 & 56.07 & 57.78 & 56.05 & 46.84 & 66.09 & 56.35 & 59.08 & 57.09  \\
    \midrule
    \multirow{4.2}{*}{GPT-3.5 Turbo} & In-context Learning & 38.69 & 60.35 & 48.36 & 54.45 & 50.46 & 40.02 & 60.60 & 46.50 & 56.28 & 50.85\\
    & Supervised Fine-tuning & 53.09 & 65.17 & 56.60 & \textbf{65.96} & 60.21 & 55.47 & \textbf{72.30} & 58.83 & 68.73 & 63.83 \\
    \addlinespace[1pt]
    \cdashline{2-12}
    \addlinespace[1pt]
    & DS\textsuperscript{2}\scalebox{1.0}[1.0]{-}ABSA\textsc{(Para)}\textsuperscript{$\ast$} & 56.86 & 64.92 & 53.15 & 61.16 & 59.02 & 60.83 & 68.41 & 54.32 & 61.87 & 61.36 \\ 
    & \textbf{DS$^2$-ABSA\textsc{(Inst)}\textsuperscript{$\ast$}} & \textbf{58.15} & \textbf{69.65} & \textbf{59.78} & 63.89 & \textbf{62.87} & \textbf{61.24} & 71.94 & \textbf{60.56} & \textbf{68.81} & \textbf{65.64} \\
    \midrule
    \multirow{3.8}{*}{Llama-3-8B-}  & In-context Learning & 41.62 & 59.84 & 46.50 & 55.01 & 50.74 & 40.70 & 61.07 & 47.13 & 56.22 & 51.28 \\

    \multirow{3.6}{*}{Instruct} & Supervised Fine-tuning & 52.52 & 63.83 & 55.65 & 61.98 & 58.50 & 56.47 & 68.85 & 59.64 & 68.11 & 63.27 \\
    \addlinespace[1pt]
    \cdashline{2-12}
    \addlinespace[1pt]
    & DS\textsuperscript{2}\scalebox{1.0}[1.0]{-}ABSA\textsc{(Para)}\textsuperscript{$\star$} & 57.29 & 65.73 & 52.68 & 60.30 & 59.00 & 60.57 & 68.83 & 54.62 & 62.44 & 61.62 \\
    & \textbf{DS$^2$-ABSA\textsc{(Inst)}\textsuperscript{$\star$}} & \textbf{60.59} & \textbf{69.94} & \textbf{58.34} & \textbf{62.90} & \textbf{62.94} & \textbf{63.21} & \textbf{72.94} & \textbf{60.51} & \textbf{68.36} & \textbf{66.26} \\
    \bottomrule
    \end{tabular}
    \caption{Main results compared with LLM-based ABSA approaches. For DS$^2$-ABSA, synthetic data generation using GPT-3.5 Turbo and Llama-3-8B-Instruct are marked with \textsuperscript{$\ast$} and \textsuperscript{$\star$}, respectively. DS$^2$-ABSA\textsc{(Para)} denotes \textsc{Paraphrase} w/ DS$^2$-ABSA, and DS$^2$-ABSA\textsc{(Inst)} denotes \textsc{Instruct}ABSA w/ DS$^2$-ABSA.}
    \label{tab: LLM}
\end{table*}

\subsection{Baselines}
To validate the effectiveness of the proposed DS$^2$-ABSA, we first select three typical ABSA models for evaluation: \textbf{TAG-BERT}~\cite{hu-etal-2019-open},  \textbf{\textsc{Paraphrase}}~\cite{zhang-etal-2021-aspect-sentiment}, and \textbf{\textsc{Instruct}ABSA}~\cite{scaria2024instructabsa}. 

We then conduct extensive comparisons with two categories of approaches. \textbf{(1) Low-resource enhancement methods}, including data augmentation: \textit{MELM}~\cite{zhou-etal-2022-melm}, \textit{AugGPT}~\cite{dai2023chataug}, \textit{CoTAM}~\cite{peng-etal-2024-controllable}; pre-training: \textit{BERT-PT}~\cite{xu2019bert}, \textit{BERT-SPT}~\cite{zhang2023empirical}, \textit{FS-ABSA}~\cite{wang2023simple}; distillation: \textit{UniNER}~\cite{zhou2024universalner}; and data synthesis: \textit{ ZeroGen}~\cite{ye-etal-2022-zerogen}, \textit{Self-Instruct}~\cite{wang-etal-2023-self-instruct}.
Comparisons with them highlight the advantages of DS$^2$-ABSA. \textbf{(2) LLM-based ABSA methods} that directly utilize LLMs to perform ABSA, such as \textit{Zero-shot Prompting}, \textit{In-context Learning}~\cite{wang2024context}, and \textit{Supervised Fine-tuning}~\cite{simmering2023large} on various backbone LLMs. Comparisons with these approaches demonstrate the potential performance gains of leveraging LLM-synthesized data to train specialized ABSA models compared to direct prompting or fine-tuning. \textbf{Detailed descriptions of these methods and more baselines are available in Appendix~\ref{sec: setting}.}

\subsection{Main Results}
\paragraph{Comparison with Low-resource Enhancement Methods.}
The results in Table~\ref{tab: slm} indicate that DS$^2$-ABSA consistently outperforms existing few-shot solutions and other LLM-oriented data generation methods. For instance, the average F1 of DS$^2$-ABSA on TAG-BERT exceeds the second-best Self-Instruct by 5.69\% and 2.99\% in the 2\%- and 5\%-shot settings, respectively. These findings underscore the superiority of our approach in low-resource scenarios, which effectively synthesize higher-quality ABSA samples. Additionally, several further observations are listed below:

(1) Original ABSA models fall short with small gold data, as exemplified by TAG-BERT achieving only 24.14\% average F1 on 2\%-shot data. Both data augmentation and synthesis methods yield considerable improvements in most situations by expanding training samples, with the latter generally surpassing the former due to the limited diversity of augmented data, particularly in 2\%-shot scenarios.


(2) Pre-training and distillation methods achieve competitive performance by leveraging large-scale external corpora that provide domain and sentiment knowledge. However, these approaches are often impractical for new domains or under privacy constraints that restrict access to such corpora. In contrast, DS$^2$-ABSA synthesizes and refines data using only small seed data, surpassing FS-ABSA and UniNER without relying on additional corpora or costly training, demonstrating both its significance and domain generalizability.

\paragraph{Comparison with LLM-based ABSA Methods.} As presented in Table~\ref{tab: LLM}, DS$^2$-ABSA outperforms LLM-based methods on nearly all datasets, confirming its superiority in low-resource settings. First, it surpasses in-context learning methods by a large margin, indicating the difficulty of aligning ABSA task requirements through direct prompting, even with GPT-4.\footnote{The specific version we use is \textit{gpt-4-0125-preview}.} Second, the proposed framework, employing either GPT-3.5 Turbo or Llama-3-8B-Instruct~\cite{dubey2024llama} for data synthesis, significantly outperforms their supervised fine-tuning counterparts with lower hardware requirements. For example, in the 5\%-shot setting, DS$^2$-ABSA\textsc{(Inst)} using these two LLMs exceeds their fine-tuning performance by 1.81\% and 2.99\%, respectively. We attribute this to LLMs retaining robust general capabilities even after fine-tuning. In comparison, DS$^2$-ABSA models are specifically optimized for ABSA, enabling them to achieve superior results. Furthermore, Llama-3-8B-Instruct and GPT-3.5 Turbo as the backbones of DS$^2$-ABSA yield similar results, demonstrating that our framework performs consistently well with LLMs of comparable capabilities.

\begin{table}[t]
    \centering
    \footnotesize
    \begin{tabular}{@{}lcccc@{}}
    \toprule
         \multirow{2.5}{*}{\textbf{Method}} & \multicolumn{2}{c}{\textbf{2\%-shot}} & \multicolumn{2}{c}{\textbf{5\%-shot}} \\
         \cmidrule(lr){2-3} \cmidrule(lr){4-5}
         & \textbf{Lap14} & \textbf{Res14} & \textbf{Lap14} & \textbf{Res14} \\
         
    \midrule
         \textbf{DS$^2$-ABSA\textsc{(Para)}} & \textbf{56.86} & \textbf{64.92} & \textbf{60.83} & \textbf{68.41} \\
    \cdashline{1-5}
    \addlinespace[2pt]
         \textit{w/o key-point-driven} & 54.98 & 60.43 & 59.05 & 66.52 \\
         \textit{w/o instance-driven} & 55.45 & 63.74 & 58.59 & 67.65 \\
         \textit{w/o normalization} & 55.08 & 63.78 & 60.20 & 67.86 \\
         \textit{w/o noise injection} & 56.51 & 64.15 & 60.61 & 67.88 \\
    \bottomrule
    \end{tabular}
    \caption{Ablation on \textsc{Paraphrase} w/ DS$^2$-ABSA.}
    \label{tab: ablation}
\end{table}

\subsection{Ablation Study}
As shown in Table~\ref{tab: ablation}, we conduct ablation experiments to assess the impact of different components in DS$^2$-ABSA. 
The observations indicate that removing any strategy of the dual-stream synthesis results in a significant performance drop, underscoring the importance of integrating both streams. Particularly, in the 2\%-shot setting, discarding key-point-driven synthesis would lead to an average F1 decrease by 3.15\%, demonstrating the importance of generating highly diverse data in scenarios with extremely limited samples. 
Additionally, removing components in the label refinement process, such as label normalization or noise injection, also leads to decreased performance. These confirm that the rule-based approach alleviates inaccuracies in synthetic data, while noisy training enhances the robustness of the student model.

\subsection{Effect of Noisy Self-training}
\begin{figure}[t]
    \centering
    \includegraphics[width=0.95\linewidth]{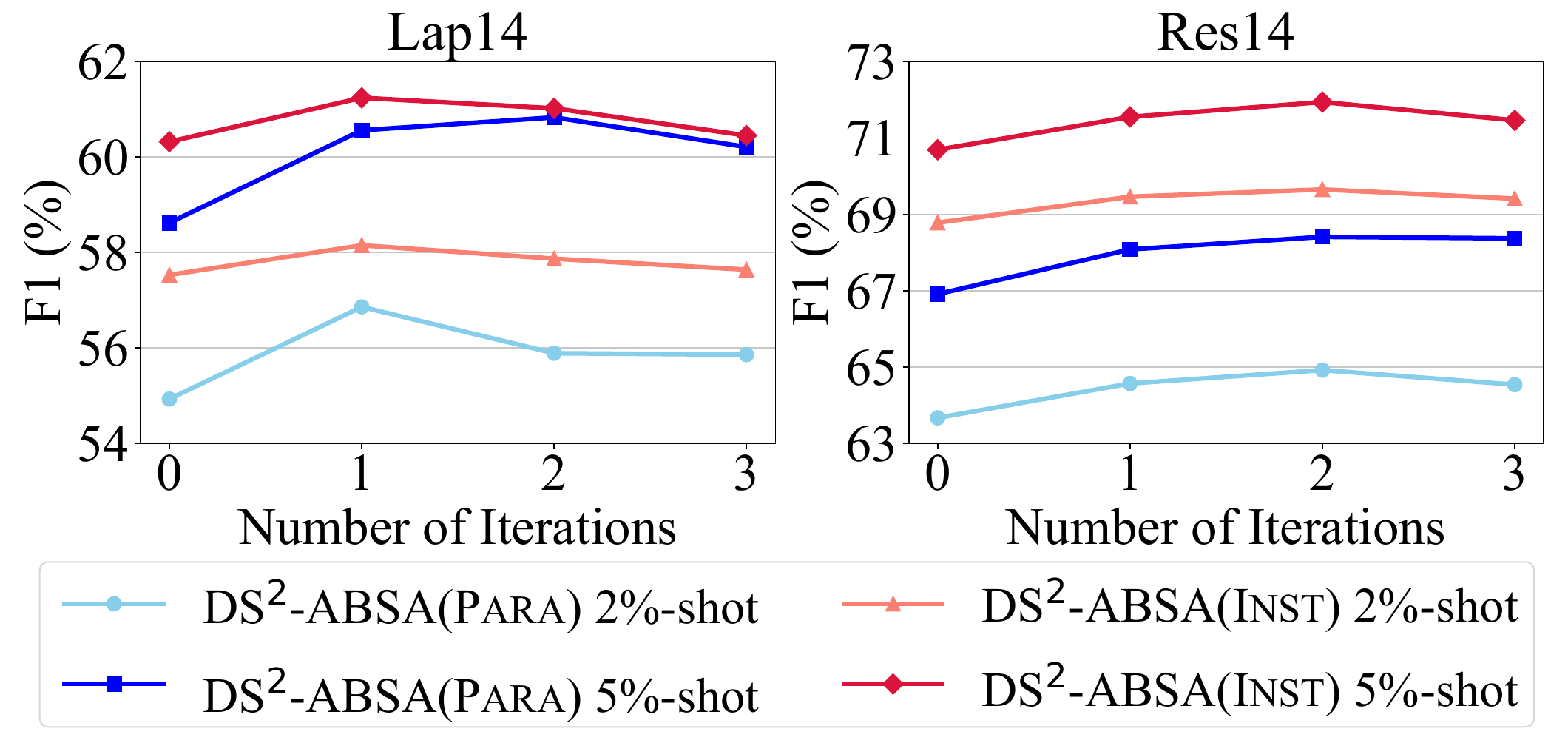}
    \caption{Effect of noisy self-training over iterations. Iteration 0 means noisy self-training is not conducted.}
    \label{fig: num_iter}
\end{figure}

To explore the impact of noisy self-training, we analyze how performances of DS$^2$-ABSA\textsc{(Para)} and DS$^2$-ABSA\textsc{(Inst)} vary with the number of iterations, with results depicted in Figure~\ref{fig: num_iter}. We observe that the F1 generally exhibits an initial increase as the number of iterations grows, confirming that self-training helps mitigate noise issues in synthetic labels.
Notably, the most significant gain occurs during the first iteration, with average F1 improvements of 1.03\% and 1.22\% under the 2\%-shot and 5\%-shot settings, respectively. Furthermore, models typically reach optimal performance after one or two iterations. We speculate that this is due to the teacher model overfitting the training data after multiple iterations, which leads to a loss of the aspect-sentiment knowledge originally provided by LLMs in the synthetic data.

\begin{figure*}[t]
    \centering
    \includegraphics[width=1.0\linewidth]{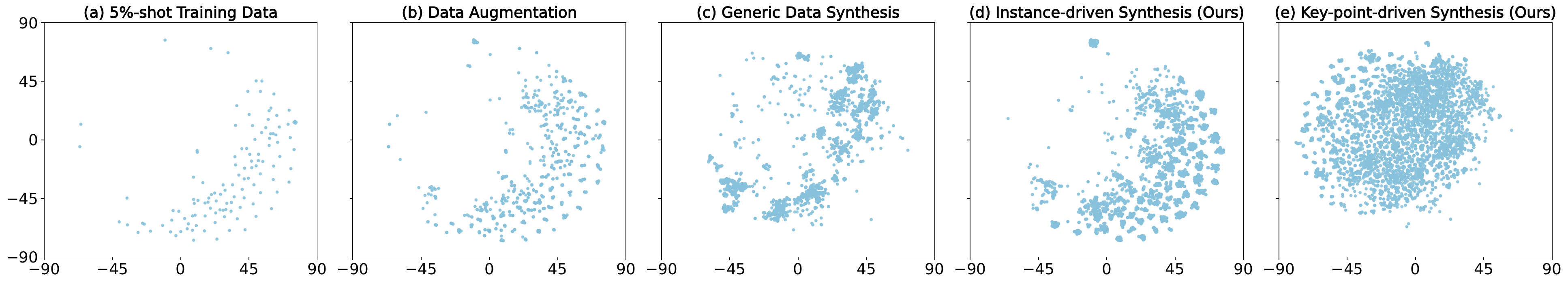}
    \caption{Data diversity comparison on Res14 under the 5\%-shot setting, including (a) few-shot gold data; (b) data augmentation (MELM, AugGPT, CoTAM); (c) generic data synthesis (ZeroGen, Self-Instruct); (d) instance-driven synthesis; and (e) key-point-driven synthesis. We use Instructor~\cite{su-etal-2023-one} for text embedding and t-SNE for visualization, displaying at most 5k samples for clarity. See Figure~\ref{fig: lap_diversity} for results on Lap14.}
    \label{fig: res_diversity}
\end{figure*}

\subsection{Exploring Diversity and Label Quality}


\paragraph{Analysis of Data Diversity.} 
As visualized in Figure~\ref{fig: res_diversity}, different data generation methods vary in their ability to enhance data diversity. Data augmentation techniques generate reviews highly similar to the original data, exhibiting poor diversity. Meanwhile, generic data synthesis slightly broadens the range of semantic embeddings. In contrast, the instance-driven technique enhances fine-grained diversity by sample merging and aspect- or context-preserved synthesis, resulting in embeddings that form larger clusters around the seed data. The key-point-driven synthesis further enriches semantic diversity, producing data with a broader and more evenly distributed range of sentence representations. Additionally, data synthesized through dual-stream methods show complementary patterns, further validating the significance of the proposed framework.

\begin{table}[t]
    \centering
    \footnotesize
    \begin{tabular}{@{}C{2.0cm}@{}C{0.812cm}@{}C{0.812cm}@{}C{0.812cm}@{}C{0.812cm}@{}C{0.812cm}@{}C{0.812cm}@{}}
    \toprule
         \multirow{2.5}{*}{\textbf{Method}} & \multicolumn{3}{c}{\textbf{Lap14}} & \multicolumn{3}{c}{\textbf{Res14}} \\
         \cmidrule(lr){2-4} \cmidrule(lr){5-7}
         & \textbf{Asp} & \textbf{Senti} & \textbf{Pair} & \textbf{Asp} & \textbf{Senti} & \textbf{Pair} \\
         
    \midrule
    \multicolumn{1}{@{}l}{Key-point-driven} & 60.08 & 64.14 & 49.60 & 57.32 & 71.42 & 50.52 \\
    \multicolumn{1}{r}{\textit{w/ label refinement}} & 68.68 & 69.64 & 63.01 & 67.99 & 81.85 & 64.94 \\
    \addlinespace[1pt]
    \cdashline{1-7}
    \addlinespace[2pt]
    \multicolumn{1}{@{}l}{Instance-driven} & 70.90 & 80.17 & 60.41 & 77.72 & 86.12 & 71.93 \\
    \multicolumn{1}{r}{\textit{w/ label refinement}} & 86.26 & 85.87 & 76.93 & 88.16 & 88.26 & 82.71 \\

    \bottomrule
    \end{tabular}
    \caption{Analysis of text-label alignment in the 5\%-shot setting using DS$^2$-ABSA\textsc{(Para)}. Asp, Senti, and, Pair represent F1 for aspects, macro-F1 for sentiments, and F1 for aspect-sentiment pairs, respectively.}

    \label{tab: consistency}
\end{table}

\paragraph{Analysis of Label Quality.} To analyze the label quality of dual-stream synthetic data and validate the effect of label refinement, we train \textsc{Instruct}ABSA with all training data and engage this model as the standard ABSA model for assessment. The results are displayed in Table~\ref{tab: consistency}. We find that instance-driven synthesis consistently achieves higher text-label consistency, significantly outperforming key-point-driven synthesis. Before refinement, the average F1 difference between them for aspect-sentiment pairs reaches 15.61\%. Furthermore, label refinement greatly enhances the alignment, with improvements of aspect-sentiment pair F1 exceeding 10\% for generated data from each stream, demonstrating its excellence.

\begin{figure}[t]
    \centering
    \includegraphics[width=0.95\linewidth]{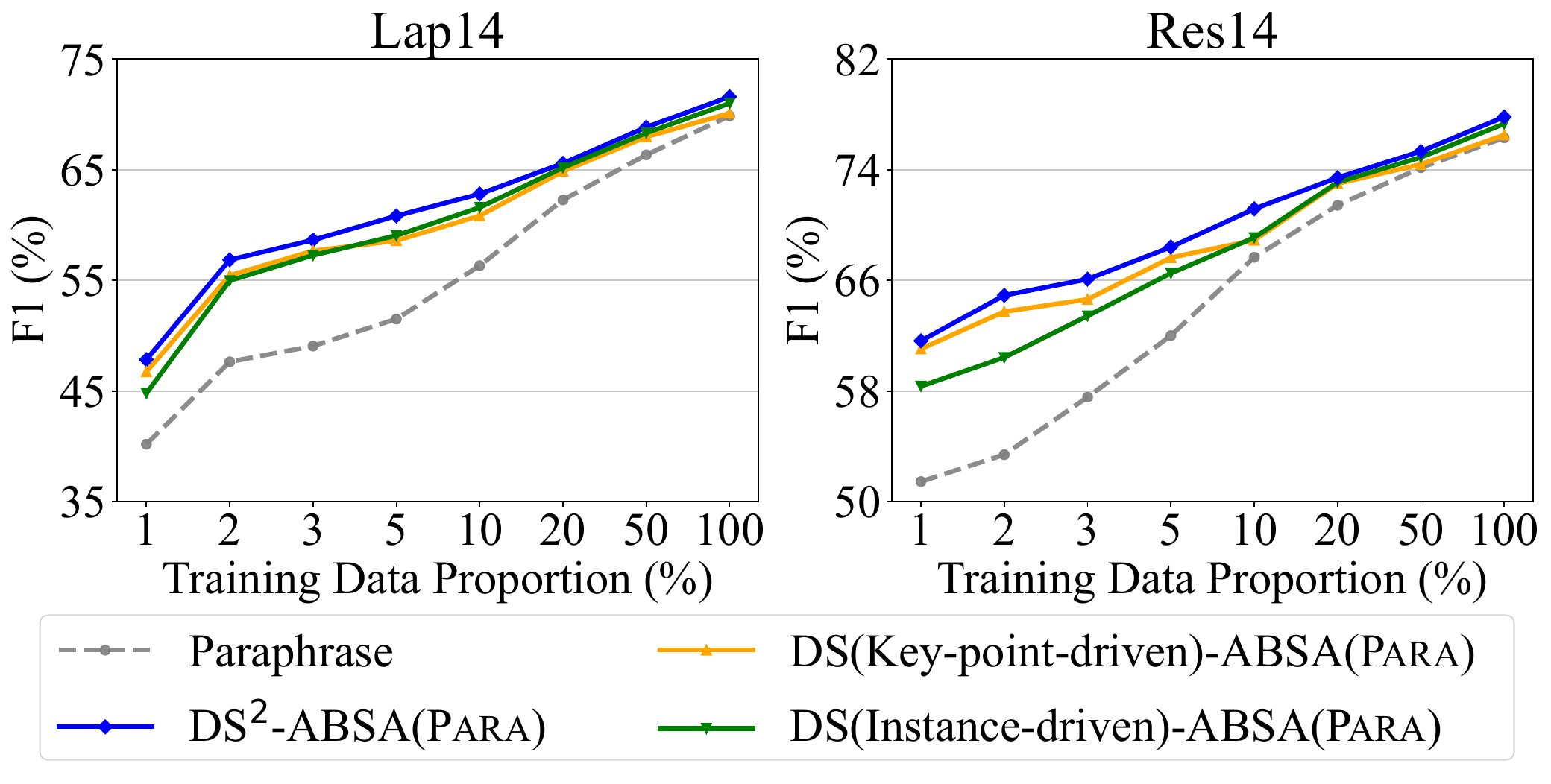}
    \caption{Effect of dual-stream synthesis methods using different training data proportions.}
    \label{fig: num_shots}
\end{figure}

\subsection{Effect of Training Data Sizes}
To investigate the effectiveness of DS$^2$-ABSA across varying training data sizes, we examine the performance of the dual-stream framework and its individual streams under different proportions of training data, as illustrated in Figure~\ref{fig: num_shots}. Initially, the original \textsc{Paraphrase} displays the worst effects, whereas DS$^2$-ABSA\textsc{(Para)} exhibits the best, proving the efficacy of dual-stream synthesis. Additionally, in conditions of scarce training data, key-point-driven synthesis outperforms instance-driven synthesis due to its ability to generate highly diverse samples via brainstormed attributes without relying on much seed data. As data volume increases, instance-driven synthesis gradually surpasses key-point-driven synthesis, suggesting that its effectiveness is positively correlated with the quantity of seed data available.

\section{Conclusion}
In this paper, we introduce DS$^2$-ABSA, a dual-stream data synthesis approach with label refinement tailored for few-shot E2E-ABSA. By leveraging LLMs with both key-point-driven and instance-driven strategies, our framework effectively generates diverse and well-aligned ABSA samples without requiring additional corpora, thereby overcoming the limitations of existing approaches. The label refinement module further enhances data quality, contributing to improved overall performance. Extensive experiments on four datasets demonstrate that DS$^2$-ABSA significantly outperforms a range of low-resource enhancement techniques and LLM-based methods, offering a promising solution for few-shot ABSA. Furthermore, we believe that our pipeline can serve as a valuable reference for data synthesis in other fields.

\section*{Limitations}
Despite the proposed DS$^2$-ABSA framework offering an effective solution for few-shot E2E-ABSA, several limitations still remain.

\begin{itemize}[leftmargin=*,itemsep=1pt]
\item For the ABSA task, annotation guidelines and specific examples for aspect term extraction and sentiment polarity identification are available, which could help improve the synthesis process. However, these have not been utilized in our current approach.
\item Although the label refinement module improves data quality, it may not fully eliminate inaccuracies arising from inherent biases in LLMs. Such biases can impair downstream performance, as residual errors in the synthetic data may lead to sub-optimal model training and predictions.
\item The prompts in key-point-driven synthesis currently rely on careful manual tuning. While this process does require an initial investment of effort, it's essential for yielding diverse reviews with promising label quality. Moreover, the consistent improvements observed across multiple domains indicate that the prompts can be effectively reused, making the cost largely a one-time effort that simplifies adaptation to new ABSA domains. 
\end{itemize}

\noindent We believe that addressing these issues provides a promising direction for further improvement.

\bibliography{custom}

\begin{thebibliography}{51}
\expandafter\ifx\csname natexlab\endcsname\relax\def\natexlab#1{#1}\fi

\bibitem[{Chan et~al.(2024)Chan, Wang, Yu, Mi, and Yu}]{chan2024scaling}
Xin Chan, Xiaoyang Wang, Dian Yu, Haitao Mi, and Dong Yu. 2024.
\newblock \href {https://arxiv.org/pdf/2406.20094} {Scaling synthetic data creation with 1,000,000,000 personas}.
\newblock \emph{arXiv preprint arXiv:2406.20094}.

\bibitem[{Dai et~al.(2023)Dai, Liu, Liao, Huang, Wu, Zhao, Liu, Liu, Li, Zhu et~al.}]{dai2023chataug}
Haixing Dai, Zhengliang Liu, Wenxiong Liao, Xiaoke Huang, Zihao Wu, Lin Zhao, Wei Liu, Ninghao Liu, Sheng Li, Dajiang Zhu, et~al. 2023.
\newblock \href {https://arxiv.org/pdf/2302.13007} {Auggpt: Leveraging chatgpt for text data augmentation}.
\newblock \emph{arXiv preprint arXiv:2302.13007}.

\bibitem[{Ding et~al.(2020)Ding, Liu, Bing, Kruengkrai, Nguyen, Joty, Si, and Miao}]{ding-etal-2020-daga}
Bosheng Ding, Linlin Liu, Lidong Bing, Canasai Kruengkrai, Thien~Hai Nguyen, Shafiq Joty, Luo Si, and Chunyan Miao. 2020.
\newblock \href {https://aclanthology.org/2020.emnlp-main.488} {{DAGA}: Data augmentation with a generation approach for low-resource tagging tasks}.
\newblock In \emph{Proceedings of the 2020 Conference on Empirical Methods in Natural Language Processing (EMNLP)}, pages 6045--6057.

\bibitem[{Dubey et~al.(2024)Dubey, Jauhri, Pandey, Kadian, Al-Dahle, Letman, Mathur, Schelten, Yang, Fan et~al.}]{dubey2024llama}
Abhimanyu Dubey, Abhinav Jauhri, Abhinav Pandey, Abhishek Kadian, Ahmad Al-Dahle, Aiesha Letman, Akhil Mathur, Alan Schelten, Amy Yang, Angela Fan, et~al. 2024.
\newblock \href {https://arxiv.org/pdf/2407.21783} {The llama 3 herd of models}.
\newblock \emph{arXiv preprint arXiv:2407.21783}.

\bibitem[{Fei et~al.(2022)Fei, Li, Li, Wu, Li, and Ji}]{ijcai2022p0572}
Hao Fei, Fei Li, Chenliang Li, Shengqiong Wu, Jingye Li, and Donghong Ji. 2022.
\newblock \href {https://doi.org/10.24963/ijcai.2022/572} {Inheriting the wisdom of predecessors: A multiplex cascade framework for unified aspect-based sentiment analysis}.
\newblock In \emph{Proceedings of the Thirty-First International Joint Conference on Artificial Intelligence, {IJCAI-22}}, pages 4121--4128.

\bibitem[{Gururangan et~al.(2020)Gururangan, Marasovi{\'c}, Swayamdipta, Lo, Beltagy, Downey, and Smith}]{gururangan-etal-2020-dont}
Suchin Gururangan, Ana Marasovi{\'c}, Swabha Swayamdipta, Kyle Lo, Iz~Beltagy, Doug Downey, and Noah~A. Smith. 2020.
\newblock \href {https://aclanthology.org/2020.acl-main.740} {Don{'}t stop pretraining: Adapt language models to domains and tasks}.
\newblock In \emph{Proceedings of the 58th Annual Meeting of the Association for Computational Linguistics}, pages 8342--8360.

\bibitem[{Hsu et~al.(2021)Hsu, Chen, Huang, and Chen}]{hsu-etal-2021-semantics}
Ting-Wei Hsu, Chung-Chi Chen, Hen-Hsen Huang, and Hsin-Hsi Chen. 2021.
\newblock \href {https://aclanthology.org/2021.emnlp-main.362} {Semantics-preserved data augmentation for aspect-based sentiment analysis}.
\newblock In \emph{Proceedings of the 2021 Conference on Empirical Methods in Natural Language Processing}, pages 4417--4422.

\bibitem[{Hu et~al.(2019)Hu, Peng, Huang, Li, and Lv}]{hu-etal-2019-open}
Minghao Hu, Yuxing Peng, Zhen Huang, Dongsheng Li, and Yiwei Lv. 2019.
\newblock \href {https://aclanthology.org/P19-1051} {Open-domain targeted sentiment analysis via span-based extraction and classification}.
\newblock In \emph{Proceedings of the 57th Annual Meeting of the Association for Computational Linguistics}.

\bibitem[{Huang et~al.(2024)Huang, Liu, Gong, Gou, Shen, Duan, and Chen}]{huang2024key}
Yiming Huang, Xiao Liu, Yeyun Gong, Zhibin Gou, Yelong Shen, Nan Duan, and Weizhu Chen. 2024.
\newblock \href {https://arxiv.org/pdf/2403.02333} {Key-point-driven data synthesis with its enhancement on mathematical reasoning}.
\newblock \emph{arXiv preprint arXiv:2403.02333}.

\bibitem[{Jiang et~al.(2023)Jiang, Drummond, and Cohn}]{jiang-etal-2023-noisy}
Fan Jiang, Tom Drummond, and Trevor Cohn. 2023.
\newblock \href {https://aclanthology.org/2023.findings-emnlp.803} {Noisy self-training with synthetic queries for dense retrieval}.
\newblock In \emph{Findings of the Association for Computational Linguistics: EMNLP 2023}, pages 11991--12008.

\bibitem[{Li et~al.(2020)Li, Chen, Quan, Ling, and Song}]{li2020conditional}
Kun Li, Chengbo Chen, Xiaojun Quan, Qing Ling, and Yan Song. 2020.
\newblock \href {https://aclanthology.org/2020.acl-main.631} {Conditional augmentation for aspect term extraction via masked sequence-to-sequence generation}.
\newblock In \emph{Proceedings of the 58th Annual Meeting of the Association for Computational Linguistics}, pages 7056--7066.

\bibitem[{Li et~al.(2023)Li, Zhu, Lu, and Yin}]{li-etal-2023-synthetic}
Zhuoyan Li, Hangxiao Zhu, Zhuoran Lu, and Ming Yin. 2023.
\newblock \href {https://aclanthology.org/2023.emnlp-main.647} {Synthetic data generation with large language models for text classification: Potential and limitations}.
\newblock In \emph{Proceedings of the 2023 Conference on Empirical Methods in Natural Language Processing}, pages 10443--10461.

\bibitem[{Liu et~al.(2023)Liu, Zhong, Ding, Jin, Du, and Tao}]{liu2023unified}
Juhua Liu, Qihuang Zhong, Liang Ding, Hua Jin, Bo~Du, and Dacheng Tao. 2023.
\newblock \href {https://ieeexplore.ieee.org/abstract/document/10184157} {Unified instance and knowledge alignment pretraining for aspect-based sentiment analysis}.
\newblock \emph{IEEE/ACM transactions on audio, speech, and language processing}, 31:2629--2642.

\bibitem[{Liu et~al.(2024)Liu, Zhou, Zhu, Chen, Bai, Xiao, and He}]{liu-etal-2024-lets}
Shunyu Liu, Jie Zhou, Qunxi Zhu, Qin Chen, Qingchun Bai, Jun Xiao, and Liang He. 2024.
\newblock \href {https://aclanthology.org/2024.lrec-main.902} {Let{'}s rectify step by step: Improving aspect-based sentiment analysis with diffusion models}.
\newblock In \emph{Proceedings of the 2024 Joint International Conference on Computational Linguistics, Language Resources and Evaluation (LREC-COLING 2024)}, pages 10324--10335.

\bibitem[{Liu et~al.(2021)Liu, Shen, and Lapata}]{liu-etal-2021-noisy}
Yang Liu, Sheng Shen, and Mirella Lapata. 2021.
\newblock \href {https://aclanthology.org/2021.naacl-main.56} {Noisy self-knowledge distillation for text summarization}.
\newblock In \emph{Proceedings of the 2021 Conference of the North American Chapter of the Association for Computational Linguistics: Human Language Technologies}, pages 692--703.

\bibitem[{Long et~al.(2024)Long, Wang, Xiao, Zhao, Ding, Chen, and Wang}]{long-etal-2024-llms}
Lin Long, Rui Wang, Ruixuan Xiao, Junbo Zhao, Xiao Ding, Gang Chen, and Haobo Wang. 2024.
\newblock \href {https://aclanthology.org/2024.findings-acl.658} {On {LLM}s-driven synthetic data generation, curation, and evaluation: A survey}.
\newblock In \emph{Findings of the Association for Computational Linguistics ACL 2024}, pages 11065--11082.

\bibitem[{OpenAI(2023)}]{openai2023gpt4}
OpenAI. 2023.
\newblock \href {http://arxiv.org/abs/2303.08774} {Gpt-4 technical report}.
\newblock \emph{CoRR, abs/2303.08774}.

\bibitem[{Ouyang et~al.(2022)Ouyang, Wu, Jiang, Almeida, Wainwright, Mishkin, Zhang, Agarwal, Slama, Ray et~al.}]{gpt35utrbo}
Long Ouyang, Jeffrey Wu, Xu~Jiang, Diogo Almeida, Carroll Wainwright, Pamela Mishkin, Chong Zhang, Sandhini Agarwal, Katarina Slama, Alex Ray, et~al. 2022.
\newblock \href {https://proceedings.neurips.cc/paper_files/paper/2022/file/b1efde53be364a73914f58805a001731-Paper-Conference.pdf} {Training language models to follow instructions with human feedback}.
\newblock \emph{Advances in neural information processing systems}, 35:27730--27744.

\bibitem[{Peng et~al.(2024)Peng, Zhang, and Shang}]{peng-etal-2024-controllable}
Letian Peng, Yuwei Zhang, and Jingbo Shang. 2024.
\newblock \href {https://aclanthology.org/2024.findings-acl.1} {Controllable data augmentation for few-shot text mining with chain-of-thought attribute manipulation}.
\newblock In \emph{Findings of the Association for Computational Linguistics: ACL 2024}, pages 1--16.

\bibitem[{Pontiki et~al.(2015)Pontiki, Galanis, Papageorgiou, Manandhar, and Androutsopoulos}]{pontiki-etal-2015-semeval}
Maria Pontiki, Dimitris Galanis, Haris Papageorgiou, Suresh Manandhar, and Ion Androutsopoulos. 2015.
\newblock \href {https://aclanthology.org/S15-2082} {{S}em{E}val-2015 task 12: Aspect based sentiment analysis}.
\newblock In \emph{Proceedings of the 9th International Workshop on Semantic Evaluation ({S}em{E}val 2015)}, pages 486--495.

\bibitem[{Pontiki et~al.(2016)Pontiki, Galanis, Papageorgiou et~al.}]{pontiki-etal-2016-semeval}
Maria Pontiki, Dimitris Galanis, Haris Papageorgiou, et~al. 2016.
\newblock \href {https://aclanthology.org/S16-1002} {{S}em{E}val-2016 task 5: Aspect based sentiment analysis}.
\newblock In \emph{Proceedings of the 10th International Workshop on Semantic Evaluation ({S}em{E}val-2016)}, pages 19--30.

\bibitem[{Pontiki et~al.(2014)Pontiki, Galanis, Pavlopoulos, Papageorgiou, Androutsopoulos, and Manandhar}]{pontiki-etal-2014-semeval}
Maria Pontiki, Dimitris Galanis, John Pavlopoulos, Harris Papageorgiou, Ion Androutsopoulos, and Suresh Manandhar. 2014.
\newblock \href {https://aclanthology.org/S14-2004} {{S}em{E}val-2014 task 4: Aspect based sentiment analysis}.
\newblock In \emph{Proceedings of the 8th International Workshop on Semantic Evaluation ({S}em{E}val 2014)}, pages 27--35.

\bibitem[{Scaria et~al.(2024)Scaria, Gupta, Goyal, Sawant, Mishra, and Baral}]{scaria2024instructabsa}
Kevin Scaria, Himanshu Gupta, Siddharth Goyal, Saurabh Sawant, Swaroop Mishra, and Chitta Baral. 2024.
\newblock \href {https://aclanthology.org/2024.naacl-short.63} {Instructabsa: Instruction learning for aspect based sentiment analysis}.
\newblock In \emph{Proceedings of the 2024 Conference of the North American Chapter of the Association for Computational Linguistics: Human Language Technologies (Volume 2: Short Papers)}, pages 720--736.

\bibitem[{Simmering and Huoviala(2023)}]{simmering2023large}
Paul~F Simmering and Paavo Huoviala. 2023.
\newblock \href {https://arxiv.org/pdf/2310.18025} {Large language models for aspect-based sentiment analysis}.
\newblock \emph{arXiv preprint arXiv:2310.18025}.

\bibitem[{Su et~al.(2023)Su, Shi, Kasai, Wang, Hu, Ostendorf, Yih, Smith, Zettlemoyer, and Yu}]{su-etal-2023-one}
Hongjin Su, Weijia Shi, Jungo Kasai, Yizhong Wang, Yushi Hu, Mari Ostendorf, Wen-tau Yih, Noah~A. Smith, Luke Zettlemoyer, and Tao Yu. 2023.
\newblock \href {https://aclanthology.org/2023.findings-acl.71} {One embedder, any task: Instruction-finetuned text embeddings}.
\newblock In \emph{Findings of the Association for Computational Linguistics: ACL 2023}, pages 1102--1121.

\bibitem[{Tian et~al.(2023{\natexlab{a}})Tian, Chen, Hu, Song, and Xia}]{tian-etal-2023-end}
Yuanhe Tian, Weidong Chen, Bo~Hu, Yan Song, and Fei Xia. 2023{\natexlab{a}}.
\newblock \href {https://aclanthology.org/2023.findings-acl.859} {End-to-end aspect-based sentiment analysis with {C}ombinatory {C}ategorial {G}rammar}.
\newblock In \emph{Findings of the Association for Computational Linguistics: ACL 2023}, pages 13597--13609.

\bibitem[{Tian et~al.(2023{\natexlab{b}})Tian, Chen, Hu, Song, and Xia}]{tian2023end}
Yuanhe Tian, Weidong Chen, Bo~Hu, Yan Song, and Fei Xia. 2023{\natexlab{b}}.
\newblock \href {https://aclanthology.org/2023.findings-acl.859} {End-to-end aspect-based sentiment analysis with combinatory categorial grammar}.
\newblock In \emph{Findings of the Association for Computational Linguistics: ACL 2023}, pages 13597--13609.

\bibitem[{Varia et~al.(2023)Varia, Wang, Halder, Vacareanu, Ballesteros, Benajiba, Anna~John, Anubhai, Muresan, and Roth}]{varia-etal-2023-instruction}
Siddharth Varia, Shuai Wang, Kishaloy Halder, Robert Vacareanu, Miguel Ballesteros, Yassine Benajiba, Neha Anna~John, Rishita Anubhai, Smaranda Muresan, and Dan Roth. 2023.
\newblock \href {https://doi.org/10.18653/v1/2023.wassa-1.3} {Instruction tuning for few-shot aspect-based sentiment analysis}.
\newblock In \emph{Proceedings of the 13th Workshop on Computational Approaches to Subjectivity, Sentiment, {\&} Social Media Analysis}, pages 19--27, Toronto, Canada. Association for Computational Linguistics.

\bibitem[{Wang et~al.(2024{\natexlab{a}})Wang, Yang, Huang, Yang, Majumder, and Wei}]{wang-etal-2024-improving-text}
Liang Wang, Nan Yang, Xiaolong Huang, Linjun Yang, Rangan Majumder, and Furu Wei. 2024{\natexlab{a}}.
\newblock \href {https://aclanthology.org/2024.acl-long.642} {Improving text embeddings with large language models}.
\newblock In \emph{Proceedings of the 62nd Annual Meeting of the Association for Computational Linguistics (Volume 1: Long Papers)}, pages 11897--11916.

\bibitem[{Wang et~al.(2024{\natexlab{b}})Wang, Ding, Luo, and Xu}]{10.1145/3626772.3657932}
Qianlong Wang, Keyang Ding, Xuan Luo, and Ruifeng Xu. 2024{\natexlab{b}}.
\newblock \href {https://doi.org/10.1145/3626772.3657932} {Improving in-context learning via sequentially selection and preference alignment for few-shot aspect-based sentiment analysis}.
\newblock In \emph{Proceedings of the 47th International ACM SIGIR Conference on Research and Development in Information Retrieval}, page 2462–2466.

\bibitem[{Wang et~al.(2024{\natexlab{c}})Wang, Xu, Ding, Liang, and Xu}]{wang2024context}
Qianlong Wang, Hongling Xu, Keyang Ding, Bin Liang, and Ruifeng Xu. 2024{\natexlab{c}}.
\newblock \href {https://aclanthology.org/2024.lrec-main.786} {In-context example retrieval from multi-perspectives for few-shot aspect-based sentiment analysis}.
\newblock In \emph{Proceedings of the 2024 Joint International Conference on Computational Linguistics, Language Resources and Evaluation (LREC-COLING 2024)}, pages 8975--8985.

\bibitem[{Wang et~al.(2023{\natexlab{a}})Wang, Kordi, Mishra, Liu, Smith, Khashabi, and Hajishirzi}]{wang-etal-2023-self-instruct}
Yizhong Wang, Yeganeh Kordi, Swaroop Mishra, Alisa Liu, Noah~A. Smith, Daniel Khashabi, and Hannaneh Hajishirzi. 2023{\natexlab{a}}.
\newblock \href {https://aclanthology.org/2023.acl-long.754} {Self-instruct: Aligning language models with self-generated instructions}.
\newblock In \emph{Proceedings of the 61st Annual Meeting of the Association for Computational Linguistics (Volume 1: Long Papers)}, pages 13484--13508.

\bibitem[{Wang et~al.(2022)Wang, Mishra, labashi et~al.}]{wang-etal-2022-super}
Yizhong Wang, Swaroop Mishra, Pegah~Alipoormo labashi, et~al. 2022.
\newblock \href {https://aclanthology.org/2022.emnlp-main.340} {Super-{N}atural{I}nstructions: Generalization via declarative instructions on 1600+ {NLP} tasks}.
\newblock In \emph{Proceedings of the 2022 Conference on Empirical Methods in Natural Language Processing}, pages 5085--5109.

\bibitem[{Wang et~al.(2024{\natexlab{d}})Wang, Xie, Feng, Ding, Yang, and Xia}]{wang2023chatgpt}
Zengzhi Wang, Qiming Xie, Yi~Feng, Zixiang Ding, Zinong Yang, and Rui Xia. 2024{\natexlab{d}}.
\newblock \href {https://openreview.net/forum?id=mUlLf50Y6H} {Is chatgpt a good sentiment analyzer? a preliminary study}.
\newblock In \emph{First Conference on Language Modeling}.

\bibitem[{Wang et~al.(2023{\natexlab{b}})Wang, Xie, and Xia}]{wang2023simple}
Zengzhi Wang, Qiming Xie, and Rui Xia. 2023{\natexlab{b}}.
\newblock \href {https://doi.org/10.1145/3539618.3591940} {A simple yet effective framework for few-shot aspect-based sentiment analysis}.
\newblock In \emph{Proceedings of the 46th International ACM SIGIR Conference on Research and Development in Information Retrieval}, pages 1765--1770.

\bibitem[{Wei and Zou(2019)}]{wei2019eda}
Jason Wei and Kai Zou. 2019.
\newblock \href {https://aclanthology.org/D19-1670} {Eda: Easy data augmentation techniques for boosting performance on text classification tasks}.
\newblock In \emph{Proceedings of the 2019 Conference on Empirical Methods in Natural Language Processing and the 9th International Joint Conference on Natural Language Processing (EMNLP-IJCNLP)}, pages 6382--6388.

\bibitem[{Xie et~al.(2020)Xie, Luong, Hovy, and Le}]{Xie_2020_CVPR}
Qizhe Xie, Minh-Thang Luong, Eduard Hovy, and Quoc~V. Le. 2020.
\newblock \href {https://openaccess.thecvf.com/content_CVPR_2020/papers/Xie_Self-Training_With_Noisy_Student_Improves_ImageNet_Classification_CVPR_2020_paper.pdf} {Self-training with noisy student improves imagenet classification}.
\newblock In \emph{Proceedings of the IEEE/CVF Conference on Computer Vision and Pattern Recognition (CVPR)}.

\bibitem[{Xu et~al.(2019)Xu, Liu, Shu, and Philip}]{xu2019bert}
Hu~Xu, Bing Liu, Lei Shu, and S~Yu Philip. 2019.
\newblock \href {https://aclanthology.org/N19-1242} {Bert post-training for review reading comprehension and aspect-based sentiment analysis}.
\newblock In \emph{Proceedings of the 2019 Conference of the North American Chapter of the Association for Computational Linguistics: Human Language Technologies, Volume 1 (Long and Short Papers)}, pages 2324--2335.

\bibitem[{Xu et~al.(2024)Xu, Cui, Yu, Kan, Shi, Zhuang, Wang, Jin, Ho, and Yang}]{xu-etal-2024-knowledge}
Ran Xu, Hejie Cui, Yue Yu, Xuan Kan, Wenqi Shi, Yuchen Zhuang, May~Dongmei Wang, Wei Jin, Joyce Ho, and Carl Yang. 2024.
\newblock \href {https://aclanthology.org/2024.findings-acl.916} {Knowledge-infused prompting: Assessing and advancing clinical text data generation with large language models}.
\newblock In \emph{Findings of the Association for Computational Linguistics ACL 2024}, pages 15496--15523.

\bibitem[{Ye et~al.(2022)Ye, Gao, Li, Xu, Feng, Wu, Yu, and Kong}]{ye-etal-2022-zerogen}
Jiacheng Ye, Jiahui Gao, Qintong Li, Hang Xu, Jiangtao Feng, Zhiyong Wu, Tao Yu, and Lingpeng Kong. 2022.
\newblock \href {https://aclanthology.org/2022.emnlp-main.801} {{Z}ero{G}en: Efficient zero-shot learning via dataset generation}.
\newblock In \emph{Proceedings of the 2022 Conference on Empirical Methods in Natural Language Processing}, pages 11653--11669.

\bibitem[{Yu et~al.(2023)Yu, Wu, Li, Bai, and Yang}]{yu2023syngen}
Chengze Yu, Taiqiang Wu, Jiayi Li, Xingyu Bai, and Yujiu Yang. 2023.
\newblock \href {https://ieeexplore.ieee.org/document/10094591} {Syngen: A syntactic plug-and-play module for generative aspect-based sentiment analysis}.
\newblock In \emph{ICASSP 2023-2023 IEEE International Conference on Acoustics, Speech and Signal Processing (ICASSP)}, pages 1--5. IEEE.

\bibitem[{Yu et~al.(2024)Yu, Jiang, Shi, YU, Liu, Zhang, Kwok, Li, Weller, and Liu}]{yu2024metamath}
Longhui Yu, Weisen Jiang, Han Shi, Jincheng YU, Zhengying Liu, Yu~Zhang, James Kwok, Zhenguo Li, Adrian Weller, and Weiyang Liu. 2024.
\newblock \href {https://openreview.net/forum?id=N8N0hgNDRt} {Metamath: Bootstrap your own mathematical questions for large language models}.
\newblock In \emph{The Twelfth International Conference on Learning Representations}.

\bibitem[{Zhang et~al.(2021)Zhang, Deng, Li, Yuan, Bing, and Lam}]{zhang-etal-2021-aspect-sentiment}
Wenxuan Zhang, Yang Deng, Xin Li, Yifei Yuan, Lidong Bing, and Wai Lam. 2021.
\newblock \href {https://aclanthology.org/2021.emnlp-main.726} {Aspect sentiment quad prediction as paraphrase generation}.
\newblock In \emph{Proceedings of the 2021 Conference on Empirical Methods in Natural Language Processing}, pages 9209--9219.

\bibitem[{Zhang et~al.(2024)Zhang, Deng, Liu, Pan, and Bing}]{zhang2024sentiment}
Wenxuan Zhang, Yue Deng, Bing Liu, Sinno Pan, and Lidong Bing. 2024.
\newblock \href {https://aclanthology.org/2024.findings-naacl.246} {Sentiment analysis in the era of large language models: A reality check}.
\newblock In \emph{Findings of the Association for Computational Linguistics: NAACL 2024}, pages 3881--3906.

\bibitem[{Zhang et~al.(2022)Zhang, Li, Deng, Bing, and Lam}]{zhang2022survey}
Wenxuan Zhang, Xin Li, Yang Deng, Lidong Bing, and Wai Lam. 2022.
\newblock \href {https://doi.org/10.1109/TKDE.2022.3230975} {A survey on aspect-based sentiment analysis: Tasks, methods, and challenges}.
\newblock \emph{IEEE Trans. on Knowl. and Data Eng.}, 35(11):11019–11038.

\bibitem[{Zhang et~al.(2023)Zhang, Yang, Liang, Chen, Qin, and Xu}]{zhang2023empirical}
Yice Zhang, Yifan Yang, Bin Liang, Shiwei Chen, Bing Qin, and Ruifeng Xu. 2023.
\newblock \href {https://aclanthology.org/2023.findings-acl.612} {An empirical study of sentiment-enhanced pre-training for aspect-based sentiment analysis}.
\newblock In \emph{Findings of the Association for Computational Linguistics: ACL 2023}, pages 9633--9651.

\bibitem[{Zhao et~al.(2024)Zhao, Jia, Viswanathan, Neubig, and Wu}]{zhao2024selfguide}
Chenyang Zhao, Xueying Jia, Vijay Viswanathan, Graham Neubig, and Tongshuang Wu. 2024.
\newblock \href {https://openreview.net/forum?id=Dt6qXZsgaU} {Self-guide: Better task-specific instruction following via self-synthetic finetuning}.
\newblock In \emph{First Conference on Language Modeling}.

\bibitem[{Zhou et~al.(2020)Zhou, Tian, Wang, Wu, Xiao, and He}]{zhou2020sentix}
Jie Zhou, Junfeng Tian, Rui Wang, Yuanbin Wu, Wenming Xiao, and Liang He. 2020.
\newblock \href {https://aclanthology.org/2020.coling-main.49} {Sentix: A sentiment-aware pre-trained model for cross-domain sentiment analysis}.
\newblock In \emph{Proceedings of the 28th international conference on computational linguistics}, pages 568--579.

\bibitem[{Zhou et~al.(2022)Zhou, Li, He, Bing, Cambria, Si, and Miao}]{zhou-etal-2022-melm}
Ran Zhou, Xin Li, Ruidan He, Lidong Bing, Erik Cambria, Luo Si, and Chunyan Miao. 2022.
\newblock \href {https://aclanthology.org/2022.acl-long.160} {{MELM}: Data augmentation with masked entity language modeling for low-resource {NER}}.
\newblock In \emph{Proceedings of the 60th Annual Meeting of the Association for Computational Linguistics (Volume 1: Long Papers)}, pages 2251--2262.

\bibitem[{Zhou et~al.(2024)Zhou, Zhang, Gu, Chen, and Poon}]{zhou2024universalner}
Wenxuan Zhou, Sheng Zhang, Yu~Gu, Muhao Chen, and Hoifung Poon. 2024.
\newblock \href {https://openreview.net/forum?id=r65xfUb76p} {Universal{NER}: Targeted distillation from large language models for open named entity recognition}.
\newblock In \emph{The Twelfth International Conference on Learning Representations}.

\bibitem[{Zhu et~al.(2024)Zhu, Zhao, Jia, and Zan}]{zhu-etal-2024-zzu}
Senbin Zhu, Hanjie Zhao, Yuxiang Jia, and Hongying Zan. 2024.
\newblock \href {https://aclanthology.org/2024.sighan-1.13} {{ZZU}-{NLP} at {SIGHAN}-2024 dim{ABSA} task: Aspect-based sentiment analysis with coarse-to-fine in-context learning}.
\newblock In \emph{Proceedings of the 10th SIGHAN Workshop on Chinese Language Processing (SIGHAN-10)}, pages 112--120.

\end{thebibliography}
\bibliographystyle{acl_natbib}

\clearpage
\newpage

\appendix

\section*{\centering\textbf{Appendix for ``DS$^2$-ABSA: Dual-Stream Data Synthesis with Label Refinement for Few-Shot Aspect-Based Sentiment Analysis''}}

We organize the appendix into three sections:

\begin{itemize}[leftmargin=*,nosep]
    \item Prompts utilized for synthetic data generation and LLM-based methods are presented in Appendix~\ref{sec: prompt};
    \item More implementation details and the descriptions of all baseline methods are presented in Appendix~\ref{appendix b};
    \item Supplemental discussions and results can be referenced in Appendix ~\ref{appendix c}.
\end{itemize}

\section{Prompt Design}
\label{sec: prompt}

For key-point-driven synthesis, we illustrate the prompts utilized in brainstorming and attribute prompting in Table~\ref{tab:bra} and~\ref{tab:attr}, respectively. For instance-driven synthesis, prompts for sample combination and selective reconstruction strategies are both presented in Table~\ref{tab:inst}.
Additionally, for LLM-based methods, including in-context learning and supervised fine-tuning, the utilized prompts are presented in Table~\ref{tab:llm_prompts}.
\section{Additional Experimental Settings}
\label{appendix b}

\subsection{Hyper-parameter Settings of Refinement} We adjust different hyper-parameters for various ABSA models to achieve optimal performance. The settings are presented in Table~\ref{tab: hyperparameters}.
\label{sec: para}

\begin{table}[h]
    \centering
    \footnotesize
    \begin{tabular}{ccccc}
    \toprule
    \textbf{Data Type} & \textbf{ABSA Model} & \textbf{BS} & \textbf{LR} & \textbf{Epochs} \\
    \midrule
    \multirow{2.2}{*}{Synthetic} 
            & TAG-BERT & 32 & 3e-6 & 5 \\
    \multirow{2}{*}{Data}
            & \textsc{Paraphrase} & 32 & 3e-5 & 5 \\
            & \textsc{InstructABSA} & 24 & 2e-5 & 5 \\
    \midrule
    \multirow{2.2}{*}{Few-shot} 
            & TAG-BERT & 8 & 5e-6 & 20 \\
    \multirow{2}{*}{Gold Data} 
            & \textsc{Paraphrase} & 8 & 1e-4 & 20 \\
            & \textsc{InstructABSA} & 8 & 1e-4 & 20 \\
    \bottomrule
    \end{tabular}
    \caption{Hyper-parameter Settings for different data types and models. BS and LR denote batch size and learning rate, respectively.}
    \label{tab: hyperparameters}
\end{table}


\subsection{Baseline Descriptions}
\label{sec: setting}

\paragraph{ABSA Models.}
(1) \textbf{TAG-BERT}~\cite{hu-etal-2019-open}: Leverages \textit{Bert-base} followed by a CRF layer for tagging. 
(2) \textbf{\textsc{Paraphrase}}~\cite{zhang-etal-2021-aspect-sentiment}: Converts ABSA into an ``aspect is sentiment'' paraphrase generation task using \textit{T5-base}.
(3) \textbf{\textsc{Instruct}ABSA}~\cite{scaria2024instructabsa}: Performs instruction-tuning with 2 positive examples, where we apply \textit{Tk-instruct-large}~\cite{wang-etal-2022-super}.

\paragraph{Data Augmentation.}
(1) \textbf{EDA}~\cite{wei2019eda}: A simple data augmentation technique that employs synonym replacement, random insertion, deletion, and swapping to enhance data variety and expand the dataset.  
(2) \textbf{CA}~\cite{li2020conditional}: Utilizes span-masking and embeds label information as conditions to generate augmented sentences with varied contextual content.  
(3) \textbf{MELM}~\cite{zhou-etal-2022-melm}: A masked language modeling approach that linearizes and embeds labels into text sequences while performing masked entity prediction to augment training data.
(4) \textbf{AugGPT}~\cite{dai2023chataug}: Utilizes LLMs to generate rephrased training samples. Each training sample is rephrased into six augmented versions, which are then combined with the original data for fine-tuning the ABSA model.
(5) \textbf{CoTAM}~\cite{peng-etal-2024-controllable}: Manipulates task-specific attributes, such as sentiment, through a three-step process: decomposition, manipulation, and reconstruction. For ABSA, it modifies the sentiment of each aspect to generate controlled augmented data.
\paragraph{Pre-training.}
(1) \textbf{BERT-PT}~\cite{xu2019bert}: Adapts pre-trained BERT to domain-specific review data through task-specific post-training. 
(2) \textbf{SentiX}~\cite{zhou2020sentix}: Proposes multi-level pre-training tasks to learn domain-invariant sentiment knowledge.
(3) \textbf{SPT-ABSA}~\cite{zhang2023empirical}: A sentiment-specific pre-training method for ABSA that integrates various sentiment knowledge from reviews. 
(4) \textbf{DAPT}~\cite{gururangan-etal-2020-dont}: Applies domain-adaptive pre-training using span corruption on 100k domain-specific reviews per domain, following~\citet{wang2023simple}.  
(5) \textbf{FS-ABSA}~\cite{wang2023simple}: Combines domain-adaptive pre-training and text-infilling fine-tuning to optimize few-shot ABSA, narrowing the gap between pre-training and downstream tasks.

\paragraph{Distillation.}
(1) \textbf{UniNER}~\cite{zhou2024universalner}: Performs targeted distillation by extracting aspect-sentiment pairs from texts using LLMs, followed by sentiment-based conversational fine-tuning to distill the knowledge into ABSA models. For a fair comparison with DS$^2$-ABSA, we collect 20k reviews for each domain, matching the data volume.

\paragraph{Data Synthesis.}
(1) \textbf{ZeroGen}~\cite{ye-etal-2022-zerogen}: Prompts LLMs to generate text by providing sentiment labels. For ABSA, we input the target sentiment into the LLM and use prompting to generate domain-specific reviews annotated with aspect-sentiment pairs.
(2) \textbf{Self-Instruct}~\cite{wang-etal-2023-self-instruct}: A typical pipeline for generating input-output pairs and performing data filtering. Here, we provide four ABSA examples to guide LLMs in generating additional reviews and labels. The synthesized data is filtered for diversity, retaining only those samples with a ROUGE-L similarity < 0.7 to existing reviews in the pool.

\paragraph{LLM-based ABSA Methods.}
(1) \textbf{Zero-shot Prompting}: Designing prompts to guide GPT-4~\cite{openai2023gpt4} for generating aspect-sentiment predictions directly without using any gold data.
(2) \textbf{In-context Learning}~\cite{wang2024context}: Retrieves in-context demonstrations for ABSA based on semantic similarity, syntactic relevance, and aspect-sentiment semantics.  
(3) \textbf{Supervised Fine-tuning}~\cite{simmering2023large}: Fine-tunes LLMs on few-shot ABSA gold data using task-specific instructions. Here, fine-tuning GPT-3.5 Turbo is implemented by Openai API.

\section{Additional Discussions}
\label{appendix c}

\subsection{Effect of ABSA Models on Full Data} 
We conduct experiments using full training data to evaluate different ABSA models, and the results are displayed in Table~\ref{tab:absa_models}. We observe that \textsc{Instruct}ABSA achieves the best performance across all datasets, demonstrating its effectiveness. Furthermore, while our synthetic data falls short compared to golden data, it still performs well in few-shot settings. Specifically, compared with the experimental results in Tabel~\ref{tab: slm}, DS$^2$-ABSA\textsc{(Inst)} can achieve 84.22\% of the full-dataset performance in average F1 using only 2\% of the training data, which increases to 87.81\% with 5\%-shot data, highlighting the quality of the synthetic data. These demonstrate the effectiveness of our method in low-resource scenarios.

\begin{table}[h]
    \centering
    \footnotesize
     \begin{tabular}{@{}C{3.2cm}@{}C{0.93cm}@{}C{0.93cm}@{}C{0.93cm}@{}C{0.93cm}@{}C{0.8cm}@{}}
        \toprule
        \textbf{Model (Params)} & \textbf{Lap14} & \textbf{Res14} & \textbf{Res15} & \textbf{Res16} & \textbf{Avg} \\
        \midrule
        TAG-BERT(109M) & 61.98 & 74.25 & 60.71 & 69.03 & 66.49 \\
        \textsc{Paraphrase}(220M) & 68.63 & 76.31 & 66.51 & 73.85 & 71.33 \\
        \textsc{Instruct}ABSA(770M) & 70.48 & 79.28 & 71.96 & 77.29 & 74.75\\
        \bottomrule
    \end{tabular}
    \caption{Full-shot performance of ABSA models.}
    \label{tab:absa_models}
\end{table}

\subsection{Effect of the Number of Synthetic Data}
\begin{figure}[h]
    \centering
    \includegraphics[width=0.95\linewidth]{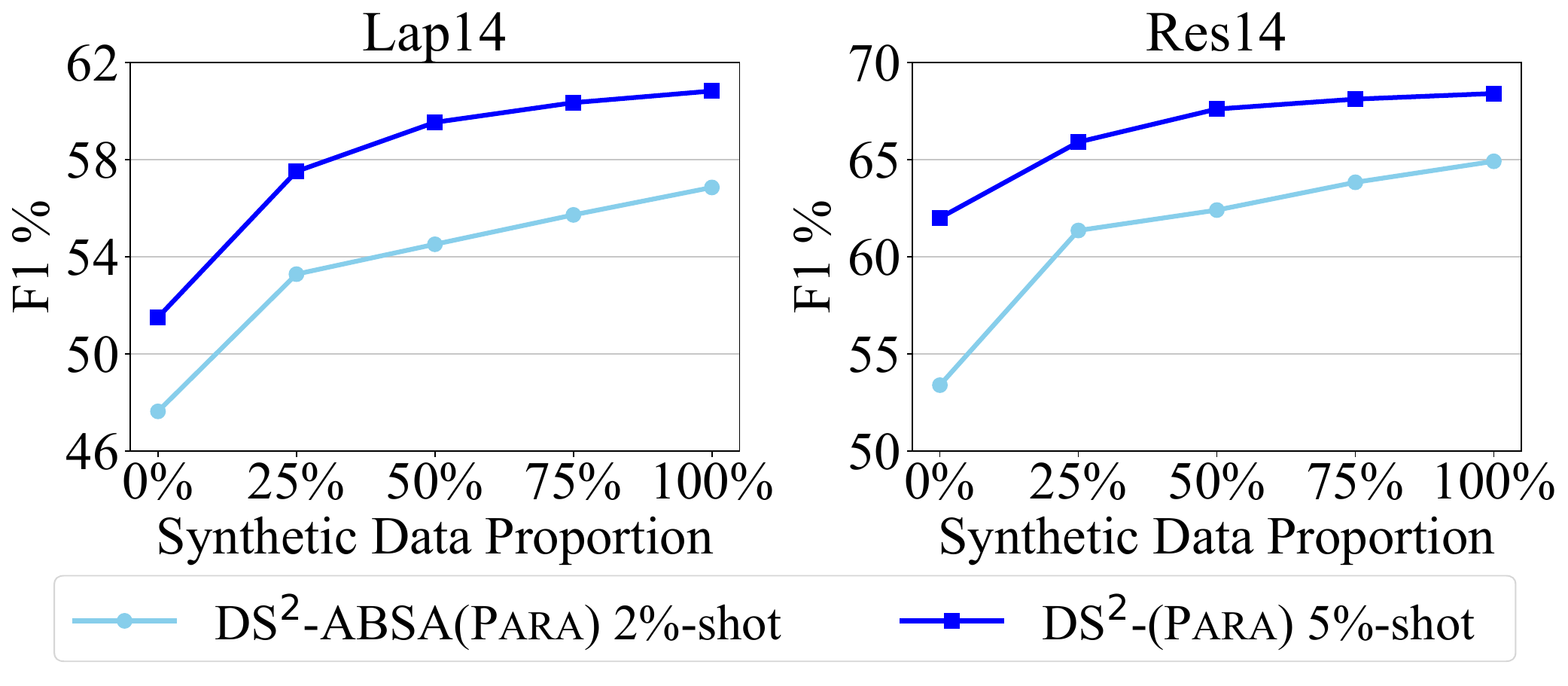}
    \caption{Impact of the number of synthetic data.}
    \label{fig: numdata}
\end{figure}

We investigate the impact of the quantity of synthetic data on model performance by sampling different proportions of synthetic data. As shown in Figure~\ref{fig: numdata}, model performance gradually improves as the amount of synthetic data increases. Additionally, we observe that in the 2\%-shot setting, changes in the quantity of synthetic data yield greater gains, with the growth rate not showing signs of decline even at 100\% synthetic data. In contrast, in the 5\%-shot setting, the growth rate gradually slows down. This indicates that when gold data is more scarce, increasing the amount of synthetic data has a greater impact on performance.

\label{sec: more_discussion}

\subsection{Data Diversity Results of Lap14}
\begin{figure*}[h]
    \centering
    \includegraphics[width=1.0\linewidth]{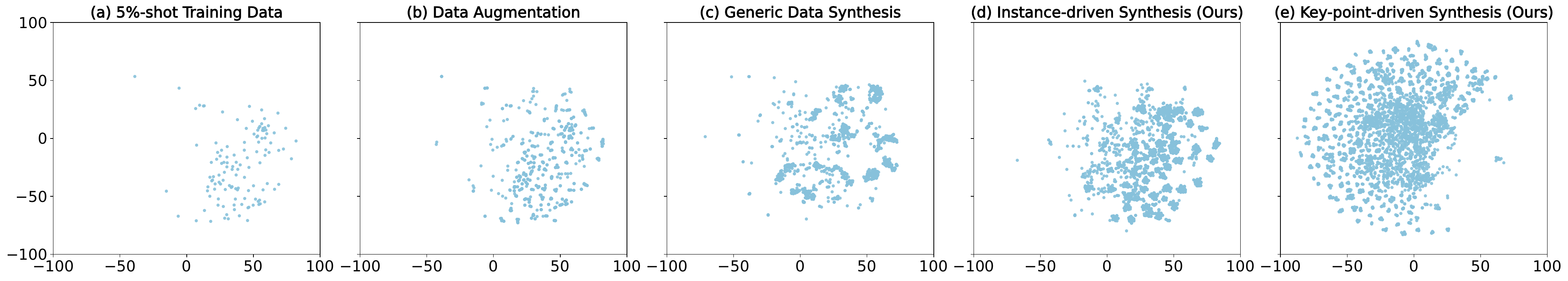}
    \caption{Data diversity comparison on Lap14 under the 5\%-shot setting, including (a) few-shot gold data; (b) data augmentation (MELM, AugGPT, CoTAM); (c) generic data synthesis (ZeroGen, Self-Instruct); (d) instance-driven synthesis; and (e) key-point-driven synthesis.}
    \label{fig: lap_diversity}
\end{figure*}

The results are illustrated in Figure~\ref{fig: lap_diversity}. We find that the diversity of different methods on Lap14 is similar to that on Res14, where key-point-driven synthesis exhibits broader diversity while semantically complementing instance-driven synthesis.

\subsection{Examples of Synthetic Data from LLMs}
To enhance clarity, Table~\ref{tab:example} and~\ref{tab:example2} present examples of data and labels synthesized by LLMs under different input conditions, which cover various strategies including attribute prompting, sample combination, and selective reconstruction.

\clearpage

\begin{table*}[h]
     \centering
    \footnotesize
    \begin{tabular}{@{}C{2.9cm}@{}p{1.8cm}@{}C{1.0cm}@{}C{1.0cm}@{}C{1.0cm}@{}C{1.0cm}@{}C{1.65cm}@{}C{1.0cm}@{}@{}C{1.0cm}@{}@{}C{1.0cm}@{}@{}C{1.0cm}@{}C{1.65cm}@{}}
    \toprule
    \multirow{2.5}{*}{\textbf{ABSA Model}} & \multirow{2.5}{*}{\textbf{Method}} & \multicolumn{5}{c}{\textbf{2\%-shot}} & \multicolumn{5}{c}{\textbf{5\%-shot}} \\
    \cmidrule(lr){3-7} \cmidrule(lr){8-12}
    & & \textbf{Lap14} & \textbf{Res14}& \textbf{Res15} & \textbf{Res16} & \textbf{Avg($\Delta$)} & \textbf{Lap14} & \textbf{Res14} & \textbf{Res15} & \textbf{Res16} & \textbf{Avg($\Delta$)} \\
    \midrule
     \multirow{4.5}{*}{TAG\scalebox{1.0}[1.0]{-}BERT} & Origin & 15.83 & 37.49 & 23.04 & 20.19 & 24.14 & 35.29 & 51.64 & 34.52 & 43.48 & 41.23 \\
     \cdashline{2-12}
     \addlinespace[1pt]
    \multirow{4}{*}{\cite{hu-etal-2019-open}} 
    & EDA\textsuperscript{$\dagger$} & 26.31 & 43.14 & 21.93 & 35.15 & 31.63\textsubscript{+7.49} & 40.77 & 52.96 & 30.06 & 45.92 & 42.43\textsubscript{+1.20} \\
    & CA\textsuperscript{$\dagger$} & 31.36 & 42.80 & 25.27 & 32.91 & 33.09\textsubscript{+8.95} & 39.14 & 53.18 & 29.22 & 46.75 & 42.07\textsubscript{+0.84} \\
    & SentiX\textsuperscript{$\ddagger$} & 36.35 & 46.07 & 28.30 & 39.80 & 37.63\textsubscript{+13.49} & 44.18 & 60.62 & 35.05 & 53.35 & 48.30\textsubscript{+7.07} \\
    & \textbf{DS\textsuperscript{2}\scalebox{1.0}[1.0]{-}ABSA}\textsuperscript{$\ast$}  & \textbf{47.30} & \textbf{60.39} & \textbf{49.49} &\textbf{59.40} & \textbf{54.15\textsubscript{+30.01}} & \textbf{47.97} & \textbf{62.37} & \textbf{49.26} & \textbf{58.40} & \textbf{54.50\textsubscript{+13.27}} \\
    \midrule
    \multirow{4.5}{*}{\textsc{Paraphrase}} & 
    Origin & 47.64 & 53.40 & 39.23  & 39.75 & 45.01 & 51.51 & 62.01 & 51.45 & 52.58 & 54.39 \\
     \cdashline{2-12}
     \addlinespace[1pt]
    \multirow{4}{*}{\cite{zhang-etal-2021-aspect-sentiment}}  
    & EDA\textsuperscript{$\dagger$} & 47.76 & 55.11 & 43.03 & 45.10 & 47.75\textsubscript{+2.74} & 51.40 & 62.14 & 49.97 & 56.41 & 54.98\textsubscript{+0.59} \\
    & CA\textsuperscript{$\dagger$} & 50.33 & 52.17 & 43.07 & 43.74 & 47.33\textsubscript{+2.32} & 51.44 & 61.63 & 49.01 & 53.55 & 53.91\textsubscript{-0.48} \\
    & DAPT\textsuperscript{$\ddagger$} & 48.03 & 56.56& 44.56 & 47.84 & 49.25\textsubscript{+4.24} & 51.81 & 63.66 & 52.37 & 58.21 & 56.51\textsubscript{+2.13} \\
    & \textbf{DS\textsuperscript{2}\scalebox{1.0}[1.0]{-}ABSA}\textsuperscript{$\ast$}  & \textbf{56.86} & \textbf{64.92} & \textbf{53.15} & \textbf{61.16} & \textbf{59.02\textsubscript{+14.01}} & \textbf{60.83} & \textbf{68.41} & \textbf{54.32} & \textbf{61.87} & \textbf{61.36\textsubscript{+6.97}} \\
    \midrule
    \multirow{4.5}{*}{\textsc{Instruct}ABSA} & Origin & 52.05 & 63.36 & 53.67 & 58.78 & 56.97 & 57.54 & 67.59 & 55.08 & 62.91 & 60.78 \\
\cdashline{2-12}
\addlinespace[1pt]
\multirow{4}{*}{\cite{scaria2024instructabsa}} 
& EDA\textsuperscript{$\dagger$} & 52.71 & 63.84 & 53.40 & 59.68 & 57.41\textsubscript{+0.44} & 56.47 & 67.06 & 53.73 & 63.29 & 60.14\textsubscript{-0.64} \\
& CA\textsuperscript{$\dagger$} & 54.39 & 62.79 & 53.28 & 57.30 & 56.93\textsubscript{-0.04} & 57.77 & 66.81 & 53.03 & 61.61 & 59.81\textsubscript{-0.97} \\
& DAPT\textsuperscript{$\ddagger$} & 55.34 & 64.90 & 55.39 & 60.46 & 59.02\textsubscript{+2.05} & 59.96 & 69.01 & 55.87 & 63.69 & 62.13\textsubscript{+1.35} \\
& \textbf{DS\textsuperscript{2}\scalebox{1.0}[1.0]{-}ABSA}\textsuperscript{$\ast$}  & \textbf{58.15} & \textbf{69.65} & \textbf{59.78} & \textbf{63.89} & \textbf{62.87\textsubscript{+5.90}} & \textbf{61.24} & \textbf{71.94} & \textbf{60.56} & \textbf{68.81} & \textbf{65.64\textsubscript{+4.86}} \\

    \bottomrule
    \end{tabular}
    \caption{More baseline results are listed here. Data augmentation, pre-training, and data synthesis methods are marked with \textsuperscript{$\dagger$}, \textsuperscript{$\ddagger$}, and \textsuperscript{$\ast$}, respectively.}
    \label{tab:more}
\end{table*}

\clearpage

\begin{table*}[t]
    \centering
    \footnotesize 
        \begin{tabular}{@{}C{2.3cm}@{}!{\vrule width 0.5pt}p{13.5cm}@{}}
            \toprule
            \textbf{Target} & \multicolumn{1}{c}{\textbf{Prompt}} \\
            \midrule
            Review Subject & \begin{minipage}{13.5cm}
                    Brainstorm a list of \textcolor{customblue}{\{domain\}} descriptions (at least 200). \\
                    \\
                    Please adhere to the following guidelines: \\
                    \phantom{ }- Names are not required. \\
                    \phantom{ }- Summarize the core features and specialties in a short, neutral sentence. \\
                     \\
                    Your output should be a Python list of strings, with each element being a description.
                \end{minipage} \\
            \midrule
            \multirow{2}{*}{Aspect Category} & \begin{minipage}{13.5cm}
                    Brainstorm a list of commonly used aspect categories in \textcolor{customblue}{\{domain\}} reviews. \\
                    \\
                    Please adhere to the following guidelines: \\
                    \phantom{ }- Aspect categories should cover various potential aspects that opinions can be expressed about within the corresponding domain. \\
                    \phantom{ }- Aspect categories are coarse-grained overviews, not including specific things. \\
                     \\
                    Your output should be a Python list of strings, with each element being a brief word denoting an aspect category. \\
                \end{minipage} \\
            \addlinespace[-4pt]
            \cdashline{2-2}
            \addlinespace[3pt]
            & Please filter the list to retain only distinct and representative aspect categories within the \textcolor{customblue}{\{domain\}} domain. Output the reason for selection along with the filtered Python list. \\
            \midrule
            Aspect Term & \begin{minipage}{13.5cm}
                Brainstorm a list of commonly used aspect terms for the aspect category \textcolor{customblue}{\{aspect category\}} within the \textcolor{customblue}{\{domain\}} domain. \\
                \\
                Please adhere to the following guidelines:\\
                \phantom{ }- Aspect terms should cover various potential things that opinions can be expressed about within the corresponding category. \\
                \phantom{ }- Aspect terms are fine-grained and concrete things. \\
                \phantom{ }- Aspect terms are single or multiword terms naming particular aspects of the target entity. \\
                \\
                Your output should be a Python list of strings, with each element being an aspect term.
                \end{minipage} \\
            \midrule
            Opinion Term & \begin{minipage}{13.5cm}
                Brainstorm a list of commonly used opinion terms for the aspect category \textcolor{customblue}{\{aspect category\}} within the \textcolor{customblue}{\{domain\}} domain. \\
                Please adhere to the following guidelines:\\
                \phantom{ }- Opinion terms refer to the expression carrying subjective emotions. \\
                \phantom{ }- Provide diverse words and phrases covering positive, negative, and neutral sentiments. \\
                \\
                Your output should be a Python list of lists, with each element being an [opinion, sentiment] pair.
                \end{minipage} \\
            \bottomrule
        \end{tabular}
        \caption{Brainstorming prompts.}
        \label{tab:bra}
\end{table*}

\begin{table*}[t]
    \centering
    \footnotesize 
        \begin{tabular}{@{}C{2.3cm}@{}!{\vrule width 0.5pt}p{13.5cm}@{}}
    \toprule
        \textbf{Method} & \multicolumn{1}{c}{\textbf{Prompt}} \\
    \midrule
        Attribute Prompting & \begin{minipage}{13.5cm}
                    Write a review sentence for the \textcolor{customblue}{\{domain\}}: \textcolor{customblue}{\{review subject\}} Label the sentence by extracting the aspect term(s) and identifying their corresponding sentiment polarity (positive, negative, or neutral). \\
                    \\
                    Requirements: \\
                    \phantom{ }- Keep a consistent \textcolor{customblue}{style} and annotation standard with the examples. \\
                    \phantom{ }- Mention the aspect term '\textcolor{customblue}{\{aspect\}}'. \\
                    \phantom{ }- Describe \textcolor{customblue}{\{aspect category\}} by the opinion term '\textcolor{customblue}{\{opinion\}}'.\\
                    \phantom{ }- Express \textcolor{customblue}{\{sentiment expression\}} across aspects. \\
                    \\
                    Here are some examples: \\
                    \phantom{ } Sentence: \textcolor{customblue}{\{sentence\}} \\
                    \phantom{ } Label: \textcolor{customblue}{\{label\}} \\
                    \phantom{ } \ldots~\ldots \\
                     \\
                     Sentence:
                \end{minipage} \\
    \bottomrule
        \end{tabular}
        \caption{Prompt template for attribute prompting.}
        \label{tab:attr}
\end{table*}

\begin{table*}[t]
    \centering
    \footnotesize 
        \begin{tabular}{@{}C{3.3cm}@{}!{\vrule width 0.5pt}p{12.5cm}@{}}
    \toprule
        \textbf{Method} & \multicolumn{1}{c}{\textbf{Prompt}} \\
    \midrule
        \multirow{12}{*}{Sample Combination} & \begin{minipage}{12.5cm}
                    Given 2 \textcolor{customblue}{\{domain\}} example reviews with the labels, please combine them to generate 4 diverse sentences. Label each sentence by extracting the aspect term(s) and determine their corresponding sentiment polarity. \\
                    \\
                    Requirements: \\
                    \phantom{ }- Keep a consistent style and annotation standard with the examples. \\
                    \phantom{ }- Maintain the same format as the example. \\
                    \phantom{ }- Combine the aspects and meanings of both examples in every generated sentence. \\
                    \\
                    Examples: \\
                    \phantom{ } Sentence: \textcolor{customblue}{\{sentence\}} \\
                    \phantom{ } Label: \textcolor{customblue}{\{label\}} \\
                    \phantom{ } Sentence: \textcolor{customblue}{\{sentence'\}} \\
                    \phantom{ } Label: \textcolor{customblue}{\{label'\}} \\
                    \\
                    4 Diverse Combined Sentences with Labels:\\
                    \\
                    1. Sentence: \\
                \end{minipage} \\
        \addlinespace[-4pt]
            \cdashline{2-2}
        \addlinespace[3pt]
         & \begin{minipage}{12.5cm}
                    Given a \textcolor{customblue}{\{domain\}} example review with the label, please paraphrase it to generate 4 diverse sentences. Label each sentence by extracting the aspect term(s) and determine their corresponding sentiment polarity. \\
                    \\
                    Requirements: \\
                    \phantom{ }- Keep a consistent style and annotation standard with the example. \\
                    \phantom{ }- Maintain the same format as the example. \\
                    \phantom{ }- The meaning of the example sentence should be unchanged. \\
                    \\
                    Example: \\
                    \phantom{ } Sentence: \textcolor{customblue}{\{sentence\}} \\
                    \phantom{ } Label: \textcolor{customblue}{\{label\}} \\
                    \\
                    4 Diverse Paraphrased Sentences with Labels:\\
                    \\
                    1. Sentence:
                \end{minipage} \\
        \addlinespace[1pt]
        \midrule
        \makecell{Selective\\Reconstruction} & \begin{minipage}{12.5cm}
                    Given a partially masked \textcolor{customblue}{\{domain\}} review sentence, please reconstruct it to generate 4 diverse sentences. Label each sentence by extracting the aspect term(s) and determine their corresponding sentiment polarity. \\
                    \\
                    Masked Sentence:  \textcolor{customblue}{\{mask sentence\}} \\
                    \\
                    Requirements: \\
                    \phantom{ }- Keep a consistent style and annotation standard with the example. \\
                    \phantom{ }- Maintain the same format as the example. \\
                    \phantom{ }- The unmasked part of the should be unchanged. \\
                    \\
                    Example: \\
                    \phantom{ } Sentence: \textcolor{customblue}{\{sentence\}} \\
                    \phantom{ } Label: \textcolor{customblue}{\{label\}} \\
                    \\
                    4 Diverse Reconstructed Sentences with Labels:\\
                    \\
                    1. Sentence:
                \end{minipage} \\
            \addlinespace[1pt]
            \bottomrule
        \end{tabular}
        \caption{Prompts for instance-driven data synthesis.}
        \label{tab:inst}
\end{table*}

\begin{table*}[t]
    \centering
    \footnotesize 
        \begin{tabular}{@{}C{3.3cm}@{}!{\vrule width 0.5pt}p{12.5cm}@{}}
    \toprule
        \textbf{Method} & \multicolumn{1}{c}{\textbf{Prompt}} \\
    \midrule
        \makecell{In-context Learning \\ \& \\ Supervised Fine-tuning}& \begin{minipage}{12.5cm}
                   Given a review, extract the aspect term(s) and determine their corresponding sentiment polarity (positive, negative, or neutral). Format the label as follows: [['aspect1', 'sentiment1'], ['aspect2', 'sentiment2'], ...]. If there are no aspect terms, use an empty list []. \textit{(Here are some examples: \ldots)} \\
                Sentence: \textcolor{customblue}{\{test input\}} \\
                Label: 
        \end{minipage} \\
        \addlinespace[1pt]
        \midrule
    \end{tabular}
    \caption{Prompts for LLM-based approaches. Note that the examples are utilized only in in-context learning.}
    \label{tab:llm_prompts}
\end{table*}

\begin{table*}
\centering
\footnotesize
    \begin{tabular}{@{}C{2.1cm}@{}!{\vrule width 0.5pt}p{13.7cm}@{}}
    \toprule
        \textbf{Method} & \multicolumn{1}{c}{\textbf{Input Prompt / Output}} \\
    \midrule
        \multirow{5}{*}{\makecell{Attribute\\Prompting}} &  \begin{minipage}{13.7cm}
        Write a review sentence for the \textcolor{customblue}{laptop}: \textcolor{customblue}{A laptop offers adaptive performance settings based on usage.}
Label the sentence by extracting the aspect term(s) and identifying their corresponding sentiment polarity (positive, negative, or neutral). \\
\\
Requirements: \\
\phantom{ }  - Keep a consistent \textcolor{customblue}{style} and annotation standard with the examples.\\
\phantom{ }  - Mention the aspect term '\textcolor{customblue}{lid rigidity}'.\\
\phantom{ }  - Describe \textcolor{customblue}{software} by the opinion term '\textcolor{customblue}{efficient}'.\\
\phantom{ } - Express \textcolor{customblue}{a consistent sentiment} across aspects. \\
\\
Here are some examples:\\
\\
Sentence: The laptop is relatively simple to use, though I bought  Macs for Dummies,  which is well worth \$2\\
Label: [['use', 'positive']]\\\\
Sentence: The computer is currently in West Verginia doe to the method of shipping choosen by Toshiba.\\
Label: [['shipping', 'negative']]\\\\
Sentence: It weighed like seven pounds or something like that.\\
Label: [['weighed', 'neutral'], ['seven pounds', 'neutral']]\\\\
Sentence: I need graphic power to run my Adobe Creative apps efficiently.\\
Label: [['graphic power', 'neutral'], ['Adobe Creative apps', 'neutral']]\\\\
Sentence: \\
\end{minipage} \\
\addlinespace[-4pt]
\cdashline{2-2}
\addlinespace[5pt]
& \begin{minipage}{13.7cm}
 The laptop offers impressive lid rigidity and efficient software for adaptive performance settings based on usage.\\
Label: [['lid rigidity', 'positive'], ['efficient software', 'positive'], ['adaptive performance settings', 'positive']]
\end{minipage} \\
\bottomrule
\end{tabular}
\caption{Examples of key-point-driven synthetic data from LLMs.}
\label{tab:example}
\end{table*}

\begin{table*}
\centering
\footnotesize
    \begin{tabular}{@{}C{2.1cm}@{}!{\vrule width 0.5pt}p{13.7cm}@{}}
    \toprule
        \textbf{Method} & \multicolumn{1}{c}{\textbf{Input Prompt / Output}} \\
    \midrule
\multirow{10}{*}{\makecell{Sample\\Combination}} & \begin{minipage}{13.7cm}
Given 2 restaurant example reviews with the labels, please combine them to generate 4 diverse sentences. Label each sentence by extracting the aspect term(s) and determine their corresponding sentiment polarity.\\
\\
Requirements:\\
\phantom{ }  - Keep a consistent style and annotation standard with the examples.\\
\phantom{ }  - Maintain the same format as the examples.\\
\phantom{ }  - Combine the aspects and meanings of both examples in every generated sentence.\\
\\
Examples:\\
\\
1.\phantom{ }Sentence: Fast service.\\
Label: [['service', 'positive']]\\
\\
2.\phantom{ }Sentence: The miso soup lacked flavor and the fish was unfortunately not as well prepared as in the past.\\
Label: [['miso soup', 'negative'], ['flavor', 'negative'], ['fish', 'negative']]\\
\\
4 Diverse Combined Sentences with Labels:\\
\\
1.\phantom{ }Sentence: \\
\end{minipage} \\
\addlinespace[-4pt]
\cdashline{2-2}
\addlinespace[5pt]
& \begin{minipage}{13.7cm}
The miso soup lacked flavor but the service was fast and efficient.\\
Label: [['miso soup', 'negative'], ['flavor', 'negative'], ['service', 'positive']]\\
\\
2.\phantom{ }Sentence: The fish was unfortunately not as well prepared as in the past, however, the service was quick and attentive.\\
Label: [['fish', 'negative'], ['service', 'positive']]\\
\\
3.\phantom{ }Sentence: Fast service made up for the lack of flavor in the miso soup.\\
Label: [['service', 'positive'], ['miso soup', 'negative'], ['flavor', 'negative']]\\
\\
4.\phantom{ }Sentence: Despite the miso soup lacking flavor, the quick service redeemed the overall dining experience.\\
Label: [['miso soup', 'negative'], ['flavor', 'negative'], ['service', 'positive']]
\end{minipage} \\
\midrule
\multirow{10}{*}{\makecell{Selective\\Reconstruction}} & \begin{minipage}{13.7cm}
Given a partially masked restaurant review sentence, please reconstruct it to generate 4 diverse sentences. Label each sentence by extracting the aspect term(s) and determine their corresponding sentiment polarity.\\
\\
Masked Sentence: <mask> anything spectacular <mask>\\
\\
Requirements:\\
\phantom{ }  - Keep a consistent style and annotation standard with the example.\\
\phantom{ }  - Maintain the same format as the example.\\
\phantom{ }  - The unmasked part of the should be unchanged.\\
\\
Example: \\
\\
Sentence: Sushi wasn't anything spectacular for the price.\\
Label: [['Sushi', 'neutral'], ['price', 'negative']]\\
\\
4 Diverse Reconstructed Sentences with Labels:\\
\\
1.\phantom{ }Sentence: \\
\end{minipage} \\
\addlinespace[-4pt]
\cdashline{2-2}
\addlinespace[5pt]
& \begin{minipage}{13.7cm}
Sushi wasn't anything spectacular for the price.\\
Label: [['Sushi', 'neutral'], ['price', 'negative']]\\
\\
2.\phantom{ }Sentence: The ambiance lacked anything spectacular, but the service was impeccable.\\
Label: [['ambiance', 'negative'], ['service', 'positive']]\\
\\
3.\phantom{ }Sentence: The dessert was anything spectacular, unlike the bland main course.\\
Label: [['dessert', 'positive'], ['main course', 'negative']]\\
\\
4.\phantom{ }Sentence: The presentation was anything spectacular, making up for the slightly high prices.\\
Label: [['presentation', 'positive'], ['prices', 'neutral']]"
\end{minipage} \\

\bottomrule
\end{tabular}
\caption{Examples of instance-driven synthetic data from LLMs.}
\label{tab:example2}
\end{table*}

\end{document}